\documentclass[twoside]{article}

\usepackage[accepted]{aistats2025}
\usepackage[round]{natbib}

\usepackage{amsmath,amsthm,amsfonts,amssymb,graphicx,bm}
\usepackage{algorithm}
\usepackage{hyperref}
\usepackage{xcolor}
\usepackage{subfig}
\usepackage{bbm}
\usepackage{enumitem}

\newcommand{\bb}{\bm{b}}

\newcommand{\bd}{\bm{d}}

\newcommand{\bh}{\bm{h}}

\newcommand{\bq}{\bm{q}}

\newcommand{\bu}{\bm{u}}
\newcommand{\bv}{\bm{v}}

\newcommand{\bx}{\bm{x}}
\newcommand{\by}{\bm{y}}
\newcommand{\bz}{\bm{z}}

\newcommand{\bA}{\bm{A}}

\newcommand{\bD}{\bm{D}}

\newcommand{\bI}{\bm{I}}

\newcommand{\bM}{\bm{M}}

\newcommand{\bP}{\bm{P}}

\newcommand{\bT}{\bm{T}}
\newcommand{\bU}{\bm{U}}

\newcommand{\bX}{\bm{X}}
\newcommand{\bY}{\bm{Y}}
\newcommand{\bZ}{\bm{Z}}

\newcommand{\cA}{\mathcal{A}}

\newcommand{\cD}{\mathcal{D}}
\newcommand{\cE}{\mathcal{E}}

\newcommand{\cG}{\mathcal{G}}
\newcommand{\cH}{\mathcal{H}}

\newcommand{\cM}{\mathcal{M}}
\newcommand{\cN}{\mathcal{N}}

\newcommand{\cP}{\mathcal{P}}

\newcommand{\cZ}{\mathcal{Z}}

\newcommand{\EE}{\mathbb{E}}
\newcommand{\GG}{\mathbb{G}}

\newcommand{\NN}{\mathbb{N}}
\newcommand{\PP}{\mathbb{P}}

\newcommand{\RR}{\mathbb{R}}

\newcommand{\ZZ}{\mathbb{Z}}

\newcommand{\beeta}{\bm{\eta}}

\newcommand{\bmu}{\bm{\mu}}

\newcommand{\bxi}{\bm{\xi}}

\newcommand{\bsigma}{\bm{\sigma}}

\newcommand{\bSigma}{\bm{\Sigma}}

\newcommand{\argmin}{\mathop{\mathrm{argmin}}}
\newcommand{\argmax}{\mathop{\mathrm{argmax}}}

\DeclareMathOperator{\var}{{\rm var}}

\DeclareMathOperator{\cov}{\rm cov}

\newcommand{\diag}{{\rm diag}}

\theoremstyle{plain} 
\newtheorem{lemma}{\textbf{Lemma}} 
\newtheorem{proposition}{\textbf{Proposition}}
\newtheorem{theorem}{\textbf{Theorem}}
\newtheorem{corollary}{\textbf{Corollary}}
\newtheorem{assumption}{\textbf{Assumption}}

 \theoremstyle{definition}
\newtheorem{definition}{\textbf{Definition}}

\newtheorem{example}{\textbf{Example}}

\makeatletter
\newenvironment{subtheorem}[1]{%
  \def\subtheoremcounter{#1}%
  \refstepcounter{#1}%
  \protected@edef\theparentnumber{\csname the#1\endcsname}%
  \setcounter{parentnumber}{\value{#1}}%
  \setcounter{#1}{0}%
  \expandafter\def\csname the#1\endcsname{\theparentnumber.\Alph{#1}}%
  \ignorespaces
}{%
  \setcounter{\subtheoremcounter}{\value{parentnumber}}%
  \ignorespacesafterend
}
\makeatother
\newcounter{parentnumber}

\newcommand{\rbr}[1]{\left(#1\right)}
\newcommand{\sbr}[1]{\left[#1\right]}
\newcommand{\cbr}[1]{\left\{#1\right\}}
\newcommand{\nbr}[1]{\left\|#1\right\|}
\newcommand{\abr}[1]{\left|#1\right|}

\newcommand{\R}{\mathbb{R}}

\newcommand{\cost}{c}
\newcommand{\vale}{v}
\newcommand{\omege}{{\bm{w}}} 
\newcommand{\ran}{{\bm{z}}}

\newcommand{\thete}{{\bm{\theta}}}

\newcommand{\boldv}{{\bm{u}}}

\newcommand{\ieo}{{\textup{IEO}}}
\newcommand{\eto}{{\textup{ETO}}}
\newcommand{\saa}{{\textup{SAA}}}
\newcommand{\kl}{{\textup{KL}}}

\newcommand{\rad}{\mathfrak{R}}
\newcommand{\radhat}{\hat{\mathfrak{R}}}
\newcommand{\tv}{\textup{TV}}

\newcommand{\mis}{\textup{mis}}
\newcommand{\st}{\textup{st}}
\newcommand{\sst}{\textup{s-st}}

\begin{document}

%
\runningtitle{Dissecting the Impact of Model Misspecification in Data-driven Optimization}

%
\runningauthor{Elmachtoub, Lam, Lan, Zhang}

\twocolumn[

\aistatstitle{Dissecting the Impact of Model Misspecification\\ in Data-Driven Optimization}

\aistatsauthor{ Adam N. Elmachtoub \And Henry Lam \And  Haixiang Lan \And Haofeng Zhang\footnotemark[1] }

\aistatsaddress{Columbia University \\ adam@ieor.columbia.edu \And Columbia University\\ henry.lam@columbia.edu \And Columbia University\\  haixiang.lan@columbia.edu
\And  Columbia University \\ Morgan Stanley\\ hz2553@columbia.edu}]

\begin{abstract}
Data-driven optimization aims to translate a machine learning model into decision-making by optimizing decisions on estimated costs. Such a pipeline can be conducted by fitting a distributional model which is then plugged into the target optimization problem. While this fitting can utilize traditional methods such as maximum likelihood, a more recent approach uses estimation-optimization integration that minimizes decision error instead of estimation error. Although intuitive, the statistical benefit of the latter approach is not well understood yet is important to guide the prescriptive usage of machine learning. In this paper, we dissect the performance comparisons between these approaches in terms of the amount of model misspecification. In particular, we show how the integrated approach offers a ``universal double benefit'' on the top two dominating terms of regret when the underlying model is misspecified, while the traditional approach can be advantageous when the model is nearly well-specified. Our comparison is powered by finite-sample tail regret bounds that are derived via new higher-order expansions of regrets and the leveraging of a recent Berry-Esseen theorem. 

\end{abstract}

\section{INTRODUCTION}

In data-driven decision-making, optimal decisions are determined by minimizing a cost function that can only be estimated from observed data. Unlike standard machine learning prediction, this calls for a \emph{prescriptive} use of data, where the goal is to obtain statistically outperforming decisions rather than predictions. To this end, a natural approach is ``estimate-then-optimize (ETO)''. Namely, we estimate unknown parameters from data via established statistical methods such as maximum likelihood estimation (MLE), and then plug into the downstream optimization task. Recently, an integrated estimation-optimization (IEO) approach has gained popularity. This approach estimates parameters by minimizing the empirical costs directly, thus accounting for the downstream optimization in the estimation procedure. Such an integrated perspective is intuitive and has propelled an array of studies in data- or machine-learning-driven optimization \citep{kao2009directed, donti2017task, elmachtoub2022smart}. Further properties and methods are studied in the integrated framework, including surrogate loss functions \citep{loke2022decision, chung2022decision}, calibration properties \citep{liu2021risk, ho2022risk}, online decision-making \citep{liu2022online}, active learning \citep{liu2023active}, differentiable optimizers \citep{berthet2020learning, blondel2020learning}, combinatorial optimization \citep{mandi2020smart, munoz2022bilevel, jeong2022exact}  and tree-based approaches \citep{elmachtoub2020decision, kallus2022stochastic}.

\footnotetext[1]{Authors are listed alphabetically.}

Despite the growing popularity, the theoretical understanding of IEO has been mostly confined to algorithmic analysis \citep{amos2017optnet,huang2024decision,bennouna2024addressing} or generalization bounds based on individual analysis as opposed to comparisons with ETO \citep{el2023generalization}. While they provide some insights on the performances of IEO, they are unable to distinguish the benefits relative to ETO, especially at an instance-specific level. An exception is the recent work \cite{elmachtoub2023estimate, hu2022fast}, which compares the regrets of IEO and ETO as data size grows. More specifically, \cite{elmachtoub2023estimate} shows that IEO has a preferable regret when the model is misspecified, and vice versa when the model is well-specified, where the preference is induced via stochastic dominance. However, because of the asymptotic nature of their analysis, the comparison characterization in \cite{elmachtoub2023estimate} is notably ``discontinuous" when transitioning from well-specified to misspecified models, in that the preference ordering of IEO and ETO changes abruptly when the model is misspecified even infinitesimally. Since in reality ``all models are wrong", the result in \cite{elmachtoub2023estimate} bears a conceptual gap with practical usage. On the other hand, \cite{hu2022fast} considers a setting when the cost function is linear, whereas we consider nonlinear cost functions.

The main goal of this paper is to provide finite-sample performance comparisons between ETO and IEO via a sharper analysis that notably captures the \emph{amount of model misspecification} in a smooth manner. Specifically, we provide a detailed dissection on the difference between the tail probabilities of ETO's and IEO's regrets, in terms of a model misspecification measure $\delta$ that, roughly speaking, signifies the suboptimality due to misspecification, as well as the tail threshold. On a high level, our results imply that IEO exhibits a lighter regret tail than ETO when the model is sufficiently misspecified, while ETO exhibits a lighter regret tail when the model is nearly, but not necessarily completely, well-specified. Importantly, both of these cases do arise in practice while the previous literature focus only on the unrealistic well-specified setting. 

Besides the above implications, our another key contribution lies in the creation of a technical roadmap that allows us to achieve the aforementioned sharp regret comparison. More precisely, we quantify the differences of regret probabilities at an accuracy that can conclude \emph{two-sided} bounds. In contrast, as we will discuss, conventional generalization analyses in machine learning based on uniformity arguments would give worst-case bounds too loose to provide comparative insights. To this end, our regret comparison roadmap consists of two novel developments. One is the derivation of higher-order expansions of IEO's and ETO's regrets. These expansions reveal a ``universal double benefit" of IEO in that the coefficients in the two most dominant terms of the IEO regret are always at most those of ETO, and these play an important role in justifying the benefit of IEO in misspecified settings. Second is the leveraging of recent Berry-Esseen bounds for $M$-estimation that allows the conversion of estimation errors into finite-sample regret bounds that sufficiently reflect the dominant error terms in the expansions. 

\section{SETUP AND PRELIMINARIES} \label{sec: uncon}
Consider a stochastic optimization problem
\begin{equation} \label{eqn: optimalsol}
\omege^* \in \argmin_{\omege\in \Omega} \left\{\vale_0(\omege):=\mathbb{E}_{P}[\cost(\omege,\ran)]     \right\}
\end{equation}
where $\omege\in \Omega\subset \mathbb{R}^p$ is the decision and $\Omega$ is an open set in $\mathbb{R}^p$, $\ran\in\cZ\subset\RR^d$ is a random vector distributed according to an unknown distribution $P\in\cP$ and $\cP$ is the collection of all potential distributions, $c(\cdot,\cdot)$ is a known cost function, and $\vale_0 (\cdot)$ is the expected cost under  $P$. 
We aim to find an optimal decision $\omege^*$.
In data-driven stochastic optimization, the true distribution $P$ is unknown, but we have i.i.d. data $\ran_1,\ldots,\ran_n$ generated from $P$. 

We use a parametric approach to infer $P$, i.e., we will estimate $P$ by a distribution in the family $\{P_\thete: \thete\in \Theta\}\subset\cP$, parameterized by $\thete$. We introduce the oracle
\begin{equation} \label{eqn: oracle}
\omege_\thete\in\argmin_{\omege\in \Omega} \{ \vale(\omege,\thete):=\EE_{P_\thete}[\cost(\omege,\ran)] \},
\end{equation}
where $\thete\in \Theta \subset\mathbb{R}^q$ is the parameter of the underlying distribution $P_\thete$ and $\Theta$ is an open set in $\mathbb{R}^q$, $\ran$ is a random vector distributed according to $P_\thete$, and $v(\cdot,\thete)$ is the expected cost under distribution $P_\thete$. Note that $\omege_\thete$ is a minimizer of problem \eqref{eqn: oracle} when $P_\thete$ is the  distribution. The true distribution $P$ may or may not be in the distribution family $\{P_\thete: \thete\in \Theta\}$. 
More precisely:
\begin{definition}[Well-Specified Model Family]\label{def: well-specidied}
The distribution family $\{P_\thete: \thete\in \Theta\}$ is \textit{well-specified} if there exists a $\thete_0\in \Theta$ such that $P=P_{\thete_0}$.     
\end{definition}

\begin{definition}[Misspecified Model Family]\label{def: misspecidied}
The distribution family $\{P_\thete: \thete\in \Theta\}$ is \textit{misspecified} if for all $\thete\in\Theta$, $P\not=P_\thete$.
\end{definition}

 We use the regret, i.e., the optimality gap or excess risk, to evaluate the quality of a decision $\omege$. 
\begin{definition}[Regret]
For any $\omege\in \Omega$, the \textit{regret} of $\omege$ is given by
$R(\omege) := \vale_0(\omege)-\vale_0(\omege^*)$,
where $\omege^*$ is an optimal solution to \eqref{eqn: optimalsol}. 
\end{definition}


\subsection{Data-Driven Optimization Approaches}
We consider two approaches to obtain a data-driven solution of \eqref{eqn: optimalsol}. Both approaches rely on using the data $\ran_1,\ldots,\ran_n$ to estimate a $\hat{\thete}$, which is then plugged into \eqref{eqn: oracle} to generate a solution $\omege_{\hat{\thete}}$.
\noindent\paragraph{Estimate-Then-Optimize (ETO):} We use maximum likelihood estimation (MLE) to infer $\thete$, i.e., 
$\hat{\thete}^{\eto}:= \arg \sup_{\thete\in\Theta}\frac{1}{n}\sum_{i=1}^n \log p_\thete (\ran_i)$
where $P_\thete$ has probability density  function $p_\thete$. By plugging $\hat{\thete}^{\eto}$ into the objective, we obtain the decision 
$\hat{\omege}^{\eto}:=\omege_{\hat{\thete}^{\eto}}=\argmin_{\omege\in \Omega} \vale(\omege,\hat{\thete}^{\eto}).$ ETO uses MLE to infer $\thete$ and then plugs $\hat{\thete}^{\eto}$ into the optimization problem \eqref{eqn: oracle} to obtain $\hat{\omege}^{\eto}$. We denote $\thete^{\kl}:= \argmax_{\thete\in \Theta} \mathbb{E}_{P} [ \log p_{\thete}(\ran)]$. This also minimizes $\kl(P,P_\thete)$ when the expected log likelihoods are finite and $\kl(P,P_{\thete_0})<\infty$ for some $\thete_0\in\Theta$. 

\noindent\paragraph{Integrated-Estimation-Optimization (IEO):} 
We infer $\thete$ by solving
\begin{align}\label{eqn: IEO loss}
    \inf_{\thete\in\Theta} \hat{\vale}_0(\omege_\thete):=\frac{1}{n}\sum_{i=1}^n\cost(\omege_\thete,\ran_i)
\end{align}
where $\omege_\thete$ is the oracle solution defined in \eqref{eqn: oracle}. The function  $c(\omege_{\thete},\ran)$ is called the \textit{IEO loss} function with respect to $\thete$. IEO integrates optimization with estimation in that the loss function used to ``train'' $\thete$ is the decision-making optimization problem evaluated on $\omege_\thete$. In other words, when we make decisions from a model parameterized by $\thete$, $\hat{\thete}^{\ieo}$ is the choice that leads to the lowest empirical risk.  By plugging $\hat{\thete}^{\ieo}$ into the objective, we obtain the decision $\hat{\omege}^{\ieo}:=\omege_{\hat{\thete}^{\ieo}}=\argmin_{\omege\in \Omega} \vale(\omege,\hat{\thete}^{\ieo}).$ We denote $\thete^*:=\argmin_{\thete\in\Theta}v_0(\omege_\thete)$.

Since the data is random, $R(\hat\omege)$ is a random variable. In this work, we shall directly compare $\eto$ and $\ieo$ by comparing the tail probabilities of the regret distributions $R(\hat\omege^{\eto})$ and $R(\hat\omege^{\ieo})$. We utilize the notion of first-order stochastic dominance \citep{quirk1962admissibility} and second-order stochastic dominance \citep{rothschild1978increasing} to rank two random variables, as defined below.

\begin{definition}[First Order Stochastic Dominance]
For any two random variables $X$, $Y$, we say that $X$ is first-order stochastically dominated by $Y$, written as $X \preceq_{\st} Y$, or $Y \succeq_{\st} X$, if 
\begin{equation}\label{def: stochasticdominance}
\PP[X>x]\le \PP[Y>x] \ \text{ for all } x\in \mathbb{R}    .
\end{equation}
\end{definition}

\begin{definition}[Second Order Stochastic Dominance]
    For any two random variables $X$ and $Y$, we say that $X$ is second-order stochastically dominated by $Y$, written as $X\preceq_{\sst} Y$, or {$Y\succeq_{\sst} X$}, if $\EE[u(X)]\leq\EE[u(Y)]$ for all non-decreasing convex function $u:\RR\to\RR$.
\end{definition}

We provide two stylized examples in data-driven decision making in our framework: the newsvendor problem and portfolio optimization.
\begin{example}
    [Multi-product Newsvendor] The multi-product newsvendor problem aims to find the optimal order quantities of $p$  products. The cost function can be represented as $\cost(\omege,\ran):=\bh^\top(\omege-\ran)^++\bb^\top(\ran-\omege)^+$, where $\ran=(z_1,\ldots,z_p)$ is the random demand each product and $\omege=(w_1,\ldots,w_p)$ is the order quantity for each product. The holding and backlogging cost are $\bh=(h_1,\ldots,h_p)$ and $\bb=(b_1,\ldots,b_p)$, respectively.
\end{example}

\begin{example}
    [Portfolio Optimization \citep{iyengar2023optimizer}] The risk-averse portfolio optimization problem with exponential utility aims to find the optimal allocation of $p$ assets. The problem has the cost function $\cost(\omege,\ran):=\exp(-\ran^\top\omege)+\gamma\|\omege\|^2$, where $\ran$ denotes the random return of each asset and $\omege$ denote the allocation weight of each asset.  
\end{example}




\textbf{Notation.} 
For nonnegative sequences $\{ a_n \}_{n=1}^{\infty}$ and $\{ b_n \}_{n=1}^{\infty}$, we write $a_n \lesssim b_n$ or $a_n = O(b_n)$ or $b_n = \Omega(a_n)$ if there exists a positive constant $C$, independent of $n$, such that $a_n \leq C b_n$. 
In addition, we write $a_n \asymp b_n$ if $a_n \lesssim b_n$ and $b_n \lesssim a_n$. 
For a general distribution $\tilde{P}$, we write $\mathbb{E}_{\tilde{P}}[\cdot]$ and $\var_{\tilde{P}}(\cdot)$ as the expectation and (co)variance with respect to the distribution $\tilde{P}$. 
Unless otherwise specified, we use $\|\cdot\|$ to denote $\|\cdot\|_2$ of a vector or a matrix. 
For a random vector $\bX$ and for any $p\geq 1$, let $\|\bX\|_p:=(\EE[\|\bX\|^p])^{1/p}$ be the $L_p$-norm of $\bX$. 
When the differentiable map $y(\boldv): \mathbb{R}^{d_1} \to \mathbb{R}$ is real-valued, the gradient $\nabla y(\boldv)$ is a row vector $1\times d_1$.
For any symmetric matrix $Q$, we write $Q\ge0$ if $Q$ is positive semi-definite and $Q> 0$ if $Q$ is positive definite. For two symmetric matrices $Q_1$ and $Q_2$, we write $Q_1\ge Q_2$ if $Q_1-Q_2\ge 0$.
Similarly, we write $Q_1> Q_2$ if $Q_1-Q_2> 0$. 
We denote $\bY_0$ as the standard multivariate Gaussian vector $N(\bm{0},\bI)$ with a proper dimension.

\begin{table*}
\caption{Summary of the comparisons between ETO and IEO.
The results are expressed in terms of the tail probability difference $\mathcal{D} = \PP( R(\hat{\omege}^{\eto})\ge t) - \PP(R(\hat{\omege}^{\ieo})\ge t)$. $\epsilon_{n,t}$ is a small constant vanishing as the sample size $n$ increases.} \label{table1}
\begin{center}
\begin{tabular}{l|l|l}
\textbf{Model Misspecification}  &$\kappa_0^\ieo < t < \kappa_0^\eto$ &$t > \kappa_0^\eto$ \\
\hline
$\delta \gg 0$    &  $\cD \ge 1 - \epsilon_{n,t}$   & $ \cD\ge - \epsilon_{n,t}$ \\
{$\delta>0$, $\tau_1>0$, $H^\eto>0$} &  $\cD \ge 1 - \epsilon_{n,t}$  & $\cD \le \epsilon_{n,t}$ (for only large $t$) \\
{$\delta=0$, $C_\mis B_0\le\tau_3$, $H^\eto>0$}     &  N/A (since $\kappa_0^\ieo=\kappa_0^\eto$)  &  $\cD \le \epsilon_{n,t}$ \\
$\delta=0$ and $B_0 = 0$  &  N/A (since $\kappa_0^\ieo=\kappa_0^\eto$)  & {$\cD \le -C_{n,t}+\epsilon_{n,t}$}   \\
\end{tabular}
\end{center}
\begingroup
Here $C_{n,t}:=\PP(n(t-\kappa_0^\ieo)\le\GG^\ieo<(1+\tau_1/\tau_6)n(t-\kappa_0^\ieo))\ge0$ in the last row. The upper bound in the second row applies when $t>\kappa_0^\ieo+(1+\tau_6/\tau_1)\delta$, under the hypotheses of Theorem \ref{thm: finitecomparemis1}.
\endgroup
\end{table*}

\section{GENERALIZATION BOUNDS}\label{sec: IEO generalization bounds}

We first derive finite-sample guarantees of IEO by utilizing conventional generalization error analysis. While this approach gives rise to regret bounds, we will also see its limitation in generating comparative insights that motivates our approach in Section \ref{sec: comparison ETO and IEO}.


In this section, we further assume the feasible region, the parameter space, and the cost function satisfy the following standard properties in learning theory.
\begin{assumption}\label{assumption: decision region and parameter space}
    The feasible decision region $\Omega$ is convex and the parameter space $\Theta$ is bounded with $E_\Theta:=\sup_{\thete\in\Theta}\|\thete\|$.
\end{assumption}
\begin{subtheorem}{assumption}\label{assumption: convexity, Lipschitzness and boundedness of cost function}
        \begin{assumption}[Convexity]\label{assumption: convexity and strong convexity of cost function} For any fixed $\ran$, $c(\cdot,\ran)$ is a convex function of $\omege$. 
For any $Q\in\cP$, $v(\omege,Q):={\EE}_Qc(\omege,\ran)$ is $\rho_c$-strongly convex  with respect to $\omege$.
    \end{assumption}
        \begin{assumption}[Lipschitzness]\label{assumption: Lipschitzness of cost function}
For any fixed $\ran$, the cost function $c(\cdot,\ran)$ is $L_c$-Lipschitz: $\forall\omege_1,\omege_2\in \Omega, \ran\in\cZ$, $|c(\omege_1,\ran)-c(\omege_2,\ran)|\leq L_c\|\omege_1-\omege_2\|$.
    \end{assumption}
        \begin{assumption}[Boundedness]\label{assumption: boundedness of cost function}
The cost function $c(\cdot,\cdot)$ is bounded: $B_c:=\sup_{\omege\in\Omega,\ran\in\cZ}c(\omege,\ran)-\inf_{\omege\in\Omega,\ran\in\cZ}c(\omege,\ran)<\infty$.
    \end{assumption}
\end{subtheorem}


We further assume the distribution family is a \textit{good parametrization}, adapted from \cite{doss2023optimal}, so that the distance between two parametric distributions are controlled by the distance between their corresponding parameters. We will give many examples that satisfy Assumption \ref{assumption: good parametrization} in Appendix \ref{sec: good parametrization examples}. 

\begin{assumption}[Good Parametrization]
    \label{assumption: good parametrization}The family of distributions $\cP_{\Theta}:=\{P_{\thete},\thete\in\Theta\}$ is a good parametrization with respect to the total variation distance $d_\tv$ if there exists $D_\Theta>0$ such that for any $\thete_1,\thete_2\in\Theta$, $d_\tv(P_{\thete_1},P_{\thete_2})\leq D_\Theta\|\thete_1-\thete_2\|$.
\end{assumption}

Under the assumptions, we have the following generalization bound:
\begin{lemma}\label{lemma: uniform convergence}
    Under Assumptions \ref{assumption: decision region and parameter space}, \ref{assumption: convexity, Lipschitzness and boundedness of cost function} and \ref{assumption: good parametrization}, there exists an absolute constant $C_\textup{abs}$ such that for any $1> \tilde{\delta}>0$, with probability at least $1-\tilde{\delta}$
    the following holds for all $\thete\in\Theta$:
\begin{align*}
v_0(\omege_\thete)\leq &\hat{v}_0(\omege_\thete)+\frac{4\sqrt{2}L_c^2C_\textup{abs}D_\Theta E_\Theta}{\rho_c}\sqrt{\frac{q}{n}}\\
&+B_c\sqrt{\frac{\log(1/\tilde{\delta})}{2n}}.
\end{align*}
\end{lemma}
Furthermore, since $\hat{\omege}^\ieo$ is an empirical minimizer of $\hat{v}_0(\cdot)$, we can bound its regret as follows.
\begin{theorem}\label{thm: IEO risk bound}
    Under Assumptions \ref{assumption: decision region and parameter space}, \ref{assumption: convexity, Lipschitzness and boundedness of cost function} and \ref{assumption: good parametrization}, there exists an absolute constant $C_\textup{abs}$ such that for any {$0<\tilde{\delta}<1$}, with probability at least $1-\tilde{\delta}$, the decision $\hat{\omege}^\ieo$ returned by IEO satisfies:
\begin{align*}
R(\hat{\omege}^\ieo)\leq &R(\omege_{\thete^*})+\frac{4\sqrt{2}L_c^2C_\textup{abs}D_\Theta E_\Theta}{\rho_c}\sqrt{\frac{q}{n}}\\
&+2B_c\sqrt{\frac{\log(2/\tilde{\delta})}{2n}}.
\end{align*}
\end{theorem}

Note that the result in Lemma \ref{lemma: uniform convergence}, which subsequently gives rise to the regret bound of IEO in Theorem \ref{thm: IEO risk bound}, holds uniformly for all $\thete\in\Theta$ including $\hat{\thete}^{\eto}$ and $\hat{\thete}^\ieo$. In other words, it is too loose to distinguish the differences between IEO and ETO. From another perspective, the attained bound has a $O(\frac{1}{\sqrt{n}})$ rate, but since we are considering parametric approaches, the {centered regret $R(\hat{\omege}^\ieo)-\kappa_0^\ieo$} is usually at a faster ${O_P}(\frac{1}{n})$ rate. Note that established fast rate analysis, even if doable for our setting, provides at best upper bounds on regrets and is still insufficient to compare ETO and IEO at an instance-specific level. 

We conclude this section by noting that \cite{el2023generalization} and \cite{qi2021integrated} also study generalization bounds for IEO of the type described above, but \cite{el2023generalization} only considers linear cost functions, and \cite{qi2021integrated} nonlinear cost but with finite discrete support on the uncertain parameter. To this end, we appear the first to analyze generalization bounds with nonlinear objectives and general parametric distributions. Nonetheless, the limitations in such bounds motivate us to consider a new analysis roadmap that we present next.


\section{DISSECTED COMPARISON}\label{sec: comparison ETO and IEO}
In this section, we present the analysis of regret comparisons between IEO and ETO, which bypasses the limitations of conventional generalization bounds. Table \ref{table1} overviews our main findings. We use two key measurements of model misspecification that control the performance comparisons between ETO and IEO, $\delta$ and $B_0$. Herein, $\delta$ is defined as $\delta:=v_0(\omege_{\thete^\kl})-v_0(\omege_{\thete^*})$, which describes loosely the distinction between the population-level ETO and IEO solutions. $B_0$ is used to characterize the distance between the true distribution and the parametric family, and its precise definition is deferred to Section \ref{sec: zeroth-order}. 




Table \ref{table1} summarizes the tail probability difference of ETO and IEO, $\cD:= \PP( R(\hat{\omege}^{\eto})\ge t) - \PP(R(\hat{\omege}^{\ieo})\ge t)$, across different scenarios of $\delta$ and $B_0$. If $\cD \le 0$, i.e., $\PP( R(\hat{\omege}^{\eto})\ge t) \le \PP(R(\hat{\omege}^{\ieo})\ge t)$, then ETO has a smaller regret tail than IEO and thus the occurrence of large regrets in ETO is less often than IEO, making ETO advantageous and vice versa.



To interpret Table \ref{table1}, we look at two cases for example. Case 1: when $\delta\gg 0$ and $\kappa_0^\ieo < t < \kappa_0^\eto$, we have
$\PP( R(\hat{\omege}^{\eto})\ge t) - \PP(R(\hat{\omege}^{\ieo})\ge t)\ge 1 - \epsilon_{n,t}$
where $\epsilon_n$ is a statistical error vanishing as $n\to\infty$. This suggests that IEO has a much lighter tail probability than ETO, and thus IEO tends to have a lower regret in this case. Case 2: When $\delta=0, B_0 \approx 0$ and $t > \kappa_0^\eto$, we have
$\PP(R(\hat{\omege}^{\eto})\ge t) - \PP( R(\hat{\omege}^{\ieo})\ge t)
\le \epsilon_{n,t}.$ In this case, ETO shows an advantage of alleviating the occurrence of extremely large regrets compared to IEO in a slightly misspecified model.

In the following, we lay out our technical developments in attaining the insights in Table \ref{table1}, which comprise two ingredients. First is high-order Taylor expansions of IEO's and ETO's regrets (Section \ref{sec: zeroth-order}). We discuss the zeroth-order ($O(1)$), first-order ($O(1/\sqrt{n})$) and second-order ($O(1/n)$) term from the expansion. These expansions reveal a universal double benefit of IEO in that its zeroth- and first-order terms are both at most those of ETO, regardless of model specification. In the second-order term, however, a universal ordering does not exist, but the ordering can be established when the model is well-specified or slightly misspecified. Our second key ingredient is the conversion of estimation errors to finite-sample regret bounds (Section \ref{sec: individualbound}) via the leveraging of recent Berry-Esseen bounds (\cite{shao2022berry}) at an accuracy that allows dissected comparisons between IEO and ETO (Section \ref{sec: finite-sample}). 



\subsection{Higher-Order Regret Expansions}\label{sec: zeroth-order}


To facilitate discussion, we define $\kappa_0^\eto:=v_0(\omege_{\thete^\kl})-v_0(\omege^*)$ and $\kappa_0^\ieo:=v_0(\omege_{\thete^*})-v_0(\omege^*)$, the regret error of the population-level ETO and IEO solutions caused by misspecification.
In addition, we denote two matrices $
{\bM}_1^\eto$, ${\bM}_1^\ieo$ as
$
{\bM}_1^\eto:=\big(\rbr{\nabla_{\thete\thete}\mathbb{E}_{P}[\log p_{\thete^{\kl}}(\ran)]}^{-1}$ $
\var_{P}(\nabla_\thete \log p_{\thete^{\kl}}(\ran))$ $(\nabla_{\thete\thete}\mathbb{E}_{P}[\log p_{\thete^{\kl}}(\ran)])^{-1}\big)^{\frac{1}{2}}$
and $\bM_1^\ieo:=\Big(\nabla_{\thete\thete}\vale_0(\omege_{\thete^*})^{-1}\var_{P}(\nabla_\thete \cost(\omege_{\thete^*},\ran))\nabla_{\thete\thete}\vale_0(\omege_{\thete^*})^{-1}\Big)^{\frac{1}{2}}$. We define two Gaussian distributions $\NN_1^\eto\overset{d}{=}\bM_1^\eto\bY_0$ and $\NN_1^\ieo\overset{d}{=}\bM_1^\ieo\bY_0$ where $\bY_0$ denotes the standard normal distribution with the corresponding dimension. The rationale for defining these notations originates from the standard asymptotic normality of M-estimation. $\bM_1^{\eto}$ (resp. $\bM_1^{\ieo}$) is the {square root of the asymptotic covariance matrix} of ETO (resp. IEO); See Proposition \ref{prop: asymptotic for all} in Appendix \ref{sec: further details in comparison}. Some well-established standard assumptions (Assumptions \ref{assumption: consistency all}, \ref{assumption: RCall}, and \ref{SCforh}) and supporting auxiliary asymptotic results are also listed in Appendix \ref{sec: further details in comparison}.

The following theorem shows the expansions of the regrets of IEO and ETO.
\begin{theorem}[Double Benefit of IEO]  \label{thm: higherorder}
Under Assumptions \ref{assumption: consistency all} and \ref{assumption: RCall}, 
we have:
     \begin{enumerate}[leftmargin=*]
    \item (Zeroth order) 
    \begin{align*} 
R(\hat{\omege}^{\eto})&\xrightarrow{P} \kappa_0^{\eto},\\
        R(\hat{\omege}^{\ieo})&\xrightarrow{P} \kappa_0^{\ieo}.
    \end{align*} 
    Moreover, $\kappa_0^{\eto}\geq\kappa_0^{\ieo}\geq 0.$
    \item (First order).
        \begin{align*}
        \sqrt{n}(R(\hat{\omege}^{\eto})-\kappa_0^\eto)&\xrightarrow{d}\nabla_\thete v_0(\omege_{\thete^\kl})\NN_1^\eto,\\
        \sqrt{n}(R(\hat{\omege}^{\ieo})-\kappa_0^\ieo)&\xrightarrow{P}0. 
    \end{align*}
Moreover, 
$0 \preceq_{\sst} \nabla_\thete v_0(\omege_{\thete^\kl})\NN_1^\eto$.  
Here $\preceq_{\sst}$ represents second-order stochastic dominance.   

\item[\text{3.}] (Second order).
        \begin{align*} 
        &n(R(\hat{\omege}^{\eto})-\kappa_0^\eto-\nabla_\thete v_0(\omege_{\thete^\kl})(\hat{\thete}^\eto-\thete^\kl))\\&\xrightarrow{d}\GG^\eto:=\frac{1}{2}{\NN_1^\eto}^\top\nabla_{\thete\thete}\vale_0(\omege_{\thete^\kl})\NN_1^\eto,\\
        &n(R(\hat{\omege}^{\ieo})-\kappa_0^\ieo)\\&\xrightarrow{d}\GG^\ieo:=\frac{1}{2}{\NN_1^\ieo}^\top\nabla_{\thete\thete}\vale_0(\omege_{\thete^*})\NN_1^\ieo.
\end{align*}
\end{enumerate}  
\end{theorem}


Theorem \ref{thm: higherorder} develops a higher-order expansion of the regret as a random variable. In particular, Theorems \ref{thm: higherorder}.1 and \ref{thm: higherorder}.2 establish a \textit{universal} ordering between ETO and IEO in terms of the zeroth- and first-order terms, in that IEO is always at least as good as ETO regardless of the used model. Therefore, {when $\|\nabla_\thete v_0(\omege_{\thete^\kl})\bM_1^\eto\|>0$,} IEO not only \emph{has a lower asymptotic regret but also a faster convergence rate compared to ETO}. In fact, the first-order regret of IEO is 0, which implies that the regret of IEO minus the model bias converges at the rate of $\frac{1}{n}$. However, the regret of ETO minus the model bias has a convergence rate $\frac{1}{\sqrt{n}}$ {in this case}. 


Unlike Theorems \ref{thm: higherorder}.1 and \ref{thm: higherorder}.2, Theorem \ref{thm: higherorder}.3 does not immediately imply a universal ordering among IEO and ETO at the second-order expansion ($\frac{1}{n}$-term). 
It is tempting to believe that the comparison of $\mathbb G^{\eto}$, $\mathbb G^{\ieo}$ in the well-specified case established in previous literature \citep{elmachtoub2023estimate} would also hold in the misspecified case. However, this is not always the case. 
The comparison in terms of the second-order expansion, $\mathbb G^{\ieo}$ vs $\mathbb G^{\eto}$, is delicate and depends on the model specification. We will zoom into the analysis of this term for the rest of this section.

We introduce some notations first.
Let $P^{\kl}=P_{\thete^{\kl}}$ and $P^{*}=P_{\thete^{*}}$ where we omit the $\thete$ notation in the distributions to avoid the confusion when we apply the gradient. For instance, 
$\nabla_{\thete\thete}\mathbb{E}_{P^{\kl}}[\log p_{\thete^{\kl}}(\ran)]=(\nabla_{\thete\thete}\mathbb{E}_{P^{\kl}}[\log p_{\thete}(\ran)])\mid_{\thete=\thete^\kl}$ 
which clearly states that the Hessian $\nabla_{\thete\thete}$ is with respect to the $\thete$ in $\log p_{\thete^{\kl}}(\ran)$, free of $P^{\kl}$. 
\begingroup
Under the reference law $P^{\kl}$, assume the displayed moments are finite, both expected Hessians are nonsingular, and the interchanges in Assumption \ref{SCforh}, including twice differentiating density normalization, hold locally with common support.
\endgroup
We denote two matrices $\tilde{\bM}_1^{\eto}$, $\tilde{\bM}_1^{\ieo}$ as
$
\tilde{\bM}_1^\eto:=\big(\rbr{\nabla_{\thete\thete}\mathbb{E}_{P^{\kl}}[\log p_{\thete^{\kl}}(\ran)]}^{-1}$ $
\var_{P^{\kl}}(\nabla_\thete \log p_{\thete^{\kl}}(\ran))$ $(\nabla_{\thete\thete}\mathbb{E}_{P^{\kl}}[\log p_{\thete^{\kl}}(\ran)])^{-1}\big)^{\frac{1}{2}}$
and $\tilde{\bM}_1^\ieo:=\Big(\nabla_{\thete\thete}\vale(\omege_{\thete^\kl},P^{\kl})^{-1}$ $\var_{P^{\kl}}(\nabla_\thete \cost(\omege_{\thete^\kl},\ran))$ $\nabla_{\thete\thete}\vale(\omege_{\thete^\kl},P^{\kl})^{-1}\Big)^{\frac{1}{2}}$.

Note that the difference between $\tilde{\bM}_1^{\ieo}$ (resp. $\tilde{\bM}_1^{\eto}$) and $\bM_1^{\ieo}$ (resp. $\bM_1^{\eto}$) is that the mean and variance are taken with respect to $P^{\kl}$ instead of the ground-truth $P$. So $\tilde{\bM}_1^{\ieo}$ (resp. $\tilde{\bM}_1^{\eto}$) is the {square root of the asymptotic covariance matrix} of IEO (resp. ETO) when $P^{\kl}$ is viewed as the ground-truth data distribution.


Next we introduce how to measure the degree of model misspecification. The most natural way is via $\kappa_0^{\eto}$ and $\kappa_0^{\ieo}$, the regret contributions of the population-level ETO and IEO solutions caused by misspecification. However, these measurements are not tight enough to trade off with data uncertainty. This points to the need of other measurements to describe a \textit{slightly misspecified} model besides $\kappa_0^{\eto}\approx \kappa_0^{\ieo}$. The intuition is that when the degree of misspecification is small, we would expect $P \approx P^{*} \approx P^{\kl}$ and $\thete^{\kl} \approx \thete^{*}$. To provide a rigorous definition,

\begin{assumption}[Degree of misspecification] \label{assumption: distributionmis} 
Suppose that the distributions $P^{\kl}$, $P^{*}$ and the true distribution $P$ satisfy that there exists ${B}_0\geq 0$ such that
\begin{align*}
    \nbr{\nabla_{\thete\thete}\mathbb{E}_{P}[\log p_{\thete^{\kl}}(\ran)] - \nabla_{\thete\thete}\mathbb{E}_{P^{\kl}}[\log p_{\thete^{\kl}}(\ran)]}&\le {B}_0 ,\\
    \nbr{\var_{P}(\nabla_\thete \log p_{\thete^{\kl}}(\ran))-\var_{P^{\kl}}(\nabla_\thete \log p_{\thete^{\kl}}(\ran))} &\le {B}_0,\\
\nbr{\nabla_{\thete\thete}\vale_0(\omege_{\thete^{*}})-\nabla_{\thete\thete}\vale(\omege_{\thete^\kl},P^{\kl})} &\le {B}_0,\\
    \nbr{\var_{P}(\nabla_\thete \cost(\omege_{\thete^*},\ran)) - \var_{P^{\kl}}(\nabla_\thete \cost(\omege_{\thete^{\kl}},\ran))} &\le {B}_0,\\
\nbr{\nabla_{\thete\thete}\vale_0(\omege_{\thete^{*}})-\nabla_{\thete\thete}\vale_0(\omege_{\thete^{\kl}})} &\le {B}_0.
\end{align*}
\end{assumption}

The assumption on the distributional similarity $P \approx P^{*} \approx P^{\kl}$ is similar to the integral probability metrics \citep{muller1997integral}, which are a well-known type of distance measurement between probability distributions. If the model is well-specified, then $P=P^*=P^\kl$ and all parameters coincide with each other. In this case, Assumption \ref{assumption: distributionmis} is satisfied with $B_0=0$. If the model is ``almost'' well-specified, from the continuity perspective, we expect the difference between the terms in the formulas in Assumption \ref{assumption: distributionmis} to be still close to $0$, say some $B_0>0$. Under Assumption \ref{assumption: distributionmis}, the next result demonstrates that the second-order regret of ETO is dominated by that of IEO up to the degree of misspecification.

\begingroup
For a symmetric matrix $B$, let $\lambda_j(B)$ denote its increasingly ordered eigenvalues. For $a\in\{\eto,\ieo\}$, define
\begin{align*}
H^\ieo&:=\nabla_{\thete\thete}v_0(\omege_{\thete^*}),\quad
H^\eto:=\nabla_{\thete\thete}v_0(\omege_{\thete^\kl}),\\
A^a&:=\bM_1^aH^a\bM_1^a,\quad
\widetilde A^a:=\tilde{\bM}_1^aH^\ieo\tilde{\bM}_1^a,\\
\bD^a&:=\operatorname{diag}(\lambda_j(A^a))_{j=1}^q,\quad
\widetilde{\bD}^a:=\operatorname{diag}(\lambda_j(\widetilde A^a))_{j=1}^q.
\end{align*}
\endgroup

\begin{theorem} \label{thm: Ginmis}
Under Assumptions \ref{assumption: distributionmis}, \ref{assumption: consistency all}, \ref{assumption: RCall}, and \ref{SCforh}, {and the reference-model conditions stated above,} there exist two random variables $Z^{\eto}$, $Z^{\ieo}$, a matrix $\Delta$ and the standard Gaussian random vector $\bm{Y}_0\sim N(\bm{0},\bI)$ such that
$Z^{\eto} \overset{d}{=} \mathbb G^{\eto},\ Z^{\ieo} \overset{d}{=} \mathbb G^{\ieo}$, and
\begingroup
\begin{align*}
Z^{\ieo}&=Z^{\eto}+\frac12\bm Y_0^\top(\bD^{\ieo}-\bD^{\eto})\bm Y_0\\
&=Z^{\eto}+\frac12\bm Y_0^\top
(\widetilde{\bD}^{\ieo}-\widetilde{\bD}^{\eto}+\Delta)\bm Y_0.
\end{align*}
\endgroup
Herein, the matrix $\Delta$ satisfies $\|\Delta\| \leq C_\mis {B}_0$, where $C_\mis$ is a problem-dependent constant, and 
\begingroup
$$\widetilde{\bD}^{\ieo}-\widetilde{\bD}^{\eto}$$
\endgroup
is positive semi-definite.
\end{theorem}

Theorem \ref{thm: Ginmis} provides a precise distance measurement between $\mathbb G^{\eto}$ and $\mathbb G^{\ieo}$ that depends on the model misspecification. Clearly, to measure the difference between $\mathbb G^{\ieo}$ and $\mathbb G^{\eto}$, we need to study the matrix {$\bD^{\ieo}-\bD^{\eto}$}. Let $\tau_1$ denote its smallest eigenvalue and $\tau_2$ denote its largest eigenvalue.

Theorem \ref{thm: Ginmis} encompasses both well-specified and misspecified cases:
   \begin{enumerate}[leftmargin=*]
\item When the model is well-specified, then we can set ${B}_0=0$, and thus $\tau_1\ge 0$ naturally holds. This immediately implies $\mathbb G^{\eto}$ is first-order stochastically dominated by $\mathbb G^{\ieo}$. 
\item When the model misspecification is small (Assumption \ref{assumption: distributionmis}), $\mathbb G^{\eto}$ is first-order stochastically dominated by $\mathbb G^{\ieo}$ with an error related to the degree of the model misspecification. This means that when the model transits from well-specified to misspecified, the relation between $\mathbb G^{\ieo}$ and $\mathbb G^{\eto}$ has  ``continuity''. We prove this rigorously in Corollary \ref{cor: Ginmis}.
\end{enumerate}  
\begin{corollary} \label{cor: Ginmis}
Under the conditions in Theorem \ref{thm: Ginmis}, let 
\begingroup
\begin{align*}
0\le\tau_3&:=\lambda_{\min}(\widetilde{\bD}^{\ieo}-\widetilde{\bD}^{\eto}).
\end{align*}
\endgroup
If ${B}_0$ satisfies ${B}_0 \le \tau_3/C_\mis$, then $\mathbb G^{\eto}$ is first-order stochastically dominated by $\mathbb G^{\ieo}$.
\end{corollary}

Corollary \ref{cor: Ginmis} indicates that for the second-order term of the regret distribution, $\mathbb G^{\eto}$ is first-order stochastically dominated by $\mathbb G^{\ieo}$ when the model misspecification is small (${B}_0$ is small). However, this fact generally does \textit{not} imply that the total regret of ETO is less than the total regret of IEO. We shall also take into account the double benefit of IEO in the first two dominating terms of regret (Theorem \ref{thm: higherorder}). It is delicate to balance these performance differences in each of the higher-order terms of the regret distribution. This will be our task in Section \ref{sec: finite-sample}. 

\subsection{Finite-Sample Bounds} \label{sec: individualbound}

In this section, we derive finite-sample regret error bounds, achieved by utilizing Berry-Esseen-type bounds on the estimation error and then converting the estimation error to the finite-sample regrets.

First, via an existing result on the finite-sample guarantee of M-estimation (Lemma \ref{lemma: berry-esseen for M-estimator} in Appendix \ref{sec: further details in comparison}), we immediately obtain the finite-sample performance of the $\thete$ solution in ETO and IEO (Proposition \ref{prop: BEforall} in Appendix \ref{sec: further details in comparison}). In the following, we integrate it with the regret analysis in Section \ref{sec: zeroth-order} to establish finite-sample performance guarantees. We introduce the following Lipschitz assumption on the Hessian of the expected cost. 

{Throughout, matrix norms are operator norms, and the constants in $C_{n,q}^a\lesssim q^{9/4}n^{-1/2}$ may be enlarged so that $C_{n,q}^a\ge q/n$ for $n,q\ge1$.}
\begin{assumption}\label{assumption: Lipschitz Hessian of v0}
We assume that $\Theta$ is {open, bounded and convex} and the function $v_0(\omege_\thete)$ {is twice continuously differentiable and} satisfies for all $\thete_1, \thete_2\in\Theta$,
\begin{align*}
\nbr{\nabla_{\thete\thete}v_0({\omege_{\thete_1}})-\nabla_{\thete\thete}v_0(\omege_{\thete_2})}\leq L_1\nbr{\thete_1-\thete_2}.
\end{align*}
\end{assumption}

From Assumption \ref{assumption: Lipschitz Hessian of v0}, it immediately follows that the Hessian function $\nabla_{\thete\thete}v_0({\omege_\thete})$ is a bounded function on $\Theta$ with $L_2:=\sup_{\thete\in\Theta}\nbr{\nabla_{\thete\thete}v_0(\omege_{\thete})}<\infty$.
\begin{subtheorem}{proposition}\label{prop: BEforallRegret}
\begin{proposition}[Finite-sample regret for ETO, part I]\label{prop: BEforsqrtnETOregret}
    Under Assumptions \ref{assumption: Lipschitz Hessian of v0}, \ref{assumption: ETOconsistency}, and \ref{assumption: RCforETO}, {if $\nbr{\nabla_\thete v_0(\omege_{\thete^\kl})\bM_1^\eto}>0$,} for all $t\in \mathbb{R}$:
\begin{align*}
&|\PP(\sqrt{n}(v_0(\hat{\omege}^\eto)-v_0(\omege_{\thete^\kl}))\geq t)\\
&-\PP(\nabla_\thete v_0(\omege_{\thete^\kl})\NN_1^\eto\geq t)|\leq G_{n,q}^\eto \\
&{\lesssim} \frac{\nbr{\bM_1^\eto}^2L_2}{\nbr{\nabla_\thete v_0(\omege_{\thete^\kl})\bM_1^\eto}} q n^{-\frac{1}{2}}{\log(en)} + C_{n,q}^\eto.
\end{align*}
Herein, ``$\lesssim$'' hides the constant independent of $n$ and $q$. Moreover, $G_{n,q}^\eto$ is independent of $t$. $C_{n,q}^\eto\lesssim q^{9/4}n^{-1/2}$ is given in Proposition \ref{prop: BEforall}.
\end{proposition}
\begin{proposition}[Finite-sample regret for IEO, part I]\label{prop: BEforsqrtnIEOregret}
Under Assumptions \ref{assumption: Lipschitz Hessian of v0}, \ref{assumption: IEOconsistency}, and \ref{assumption: RCforIEO}, for all $t>0$:
\begin{align*}
        &\PP(\sqrt{n}(v_0(\hat{\omege}^\ieo)-v_0(\omege_{\thete^*}))\geq t)\\
        &\le G_{n,q,t}^\ieo{\lesssim} q\exp\rbr{-\frac{\sqrt{n}t}{qL_2\nbr{\bM_1^\ieo}^2}}+C_{n,q}^\ieo.
    \end{align*}
Herein, ``$\lesssim$'' hides the constant independent of $n$ and $q$. $C_{n,q}^\ieo\lesssim q^{9/4}n^{-1/2}$ is given in Proposition \ref{prop: BEforall}. 
\end{proposition}
\end{subtheorem}

Proposition \ref{prop: BEforallRegret} indicates the finite-sample regret bounds for ETO and IEO with respect to the first-order term $O(\frac{1}{\sqrt{n}})$. 
Importantly, it shows that the finite-sample regret error vanishes at the rate (essentially) $O(\frac{1}{\sqrt{n}})$, which is the same as the finite-sample estimation error (Proposition \ref{prop: BEforall}). There is an additional $(\log n)$ in the bound of ETO since the first-order regret of ETO is non-zero (Theorem \ref{thm: higherorder}), which introduces an additional small error.

The next proposition indicates the finite-sample regret bounds with respect to the second-order term $O(\frac{1}{n})$.

\begin{subtheorem}{proposition}\label{prop: BEforallRegret2}
\begin{proposition}[Finite-sample regret for ETO, part II]\label{prop: BEfornETOregret}
Under Assumptions \ref{assumption: Lipschitz Hessian of v0}, \ref{assumption: ETOconsistency}, and \ref{assumption: RCforETO}, and assume that $\nabla_{\thete\thete}v_0(\omege_{\thete^\kl}){>0}$, we have for all $t>0$,
\begin{align*}
&|\PP(n(v_0(\hat{\omege}^\eto)-v_0(\omege_{\thete^\kl})-\nabla_\thete v_0(\omege_{\thete^\kl})(\hat{\thete}^\eto-\thete^\kl))\\&\geq t)-\PP(\GG^{\eto}\geq t)|{\le} D_{n,q}^\eto.
\end{align*}
When $q=1$,
$$D_{n,1}^\eto \lesssim \frac{1}{\sqrt{\lambda_1}} L_1^{\frac{1}{2}} \nbr{\bM_1^\eto}^{\frac{3}{2}} ({\log(en)})^{\frac{3}{4}} n^{-\frac{1}{4}} + C_{n,1}^\eto$$
and when $q\ge 2$,
\begin{align*}
&D_{n,q}^\eto \lesssim C_{n,q}^\eto + \\
&\rbr{\rbr{\sum_{i=1}^q\lambda_i^2}\rbr{\sum_{i=2}^q\lambda_i^2}}^{-\frac{1}{4}}L_1\nbr{\bM_1^\eto}^3q^{3/2}({\log(en)})^{\frac{3}{2}} n^{-\frac{1}{2}}    
\end{align*}
Herein, ``$\lesssim$'' hides constants independent of $n$ and $q$. $D_{n,q}^\eto$ is independent of $t$. $C_{n,q}^\eto\lesssim q^{9/4}n^{-1/2}$ is given in Proposition \ref{prop: BEforall}. ${\lambda_1\ge\cdots\ge\lambda_q>0}$ are the eigenvalues of $\frac{1}{2}{\bM}_1^{\eto}{\nabla_{\thete\thete}}v_0(\omege_{\thete^\kl}){\bM}_1^{\eto}$.
\end{proposition}
\begin{proposition}[Finite-sample regret for IEO, part II]\label{prop: BEfornIEOregret}
    Under Assumptions \ref{assumption: Lipschitz Hessian of v0}, \ref{assumption: IEOconsistency}, and \ref{assumption: RCforIEO}, for all $t>0$, 
        \begin{align*}
 \abr{\PP(nv_0(\hat{\omege}^\ieo)-nv_0(\omege_{\thete^*})\geq t)-\PP(\GG^{\ieo}\geq t)}\le D_{n,q}^\ieo.
    \end{align*}
When $q=1$,
$$D_{n,1}^\ieo \lesssim \frac{1}{\sqrt{\lambda_1}} L_1^{\frac{1}{2}} \nbr{\bM_1^\ieo}^{\frac{3}{2}} ({\log(en)})^{\frac{3}{4}} n^{-\frac{1}{4}} + C_{n,1}^\ieo$$
and when $q\ge 2$,
\begin{align*}
&D_{n,q}^\ieo \lesssim C_{n,q}^\ieo + \\
&\rbr{\rbr{\sum_{i=1}^q\lambda_i^2}\rbr{\sum_{i=2}^q\lambda_i^2}}^{-\frac{1}{4}}L_1\nbr{\bM_1^\ieo}^3q^{3/2}({\log(en)})^{\frac{3}{2}} n^{-\frac{1}{2}}    
\end{align*}
Herein, ``$\lesssim$'' hides constants independent of $n$ and $q$. $D_{n,q}^\ieo$ is independent of $t$. $C_{n,q}^\ieo\lesssim q^{9/4}n^{-1/2}$ is given in Proposition \ref{prop: BEforall}. ${\lambda_1\ge\cdots\ge\lambda_q>0}$ are the eigenvalues of $\frac{1}{2}{\bM}_1^{\ieo}{\nabla_{\thete\thete}}v_0(\omege_{\thete^*}){\bM}_1^{\ieo}$.
\end{proposition}
\end{subtheorem}

Propositions \ref{prop: BEforallRegret} and \ref{prop: BEforallRegret2} formally establish the individual finite-sample regret bounds for ETO and IEO by converting Berry–Esseen bounds for the estimation error to finite-sample regrets, using the higher-order expansion result we develop in Section \ref{sec: zeroth-order}. Unlike generalization bound in Theorem \ref{thm: IEO risk bound} (which are upper bounds), our bounds consists of both upper and lower bounds and thus provide a tighter characterization of the finite-sample regret distribution. These results will be used to derive the ultimate regret comparisons in Section \ref{sec: finite-sample}.

An immediate implication of the above theorems is a high probability bound of the IEO regret and its fast rate compared to previous generalization bounds. 
\begin{corollary}
\label{cor: Fast rate, Berry Esseen}
Under Assumptions \ref{assumption: Lipschitz Hessian of v0}, \ref{assumption: IEOconsistency}, and \ref{assumption: RCforIEO}, there exists a problem dependent $C_\textup{prob}$, such that for any ${0<\varepsilon<1}$, when $n$ satisfies $C_{\textup{prob}}({\log(en)})^{\frac{3}{2}} n^{-\frac{1}{2}}\leq \varepsilon/2$ (for $q\ge 2$) or $C_{\textup{prob}}({\log(en)})^{\frac{3}{4}} n^{-\frac{1}{4}} \leq \varepsilon/2$ (for $q=1$), with probability at least $1-\varepsilon$, $R(\hat{\omege}^\ieo)\leq \kappa_0^\ieo+\frac{F_{\GG^{\ieo}}^{-1}(1-\varepsilon/2)}{n}$.
\end{corollary}
The above corollary implies when the sample size $n$ is larger than an explicit threshold, the convergence rate of {the centered regret $R(\hat{\omege}^\ieo)-\kappa_0^\ieo$} is ${O_P}(1/n)$, matching the results in the asymptotic analysis.

\subsection{Comparisons on Finite-Sample Regret} \label{sec: finite-sample}
With the above developments, we now derive regret comparisons that draw our insights in Table \ref{table1}. Our main results consist of two theorems that formally summarize the comparison in terms of the tail probability of the regret distribution, one in a generally misspecified model and one in a ``slightly misspecified'' model (including the well-specified model). The assumptions in the two theorems are not mutually exclusive.

Recall $\delta=\kappa_0^\eto-\kappa_0^\ieo$. In Section \ref{sec: zeroth-order}, we have established the “double benefit” property of IEO. Intuitively, when $\delta$ is relatively large or the sample size $n$ is relatively large, the major component of the regret distribution would be the first two order terms of the regret where IEO is strictly better.
Combining this observation with the finite-sample regret bounds in Section \ref{sec: individualbound}, we can build an explicit finite-sample regret comparison.

\begin{theorem} [Lower Bound on $\cD$]\label{thm: finitecomparemis2}
Suppose Assumptions \ref{assumption: distributionmis}, \ref{assumption: Lipschitz Hessian of v0}, \ref{assumption: consistency all}, \ref{assumption: RCall}, and \ref{SCforh} hold. {Assume $\|\nabla_\thete v_0(\omege_{\thete^\kl})\bM_1^\eto\|>0$.} We have:

Case 1: $t\le \kappa_0^{\ieo}:$
\begin{align*}
\PP(R(\hat{\omege}^{\eto})\ge t) - \PP( R(\hat{\omege}^{\ieo})\ge t) =0
\end{align*}

Case 2: {$t > \kappa_0^\ieo$:}
\begin{align*}
\PP(R(\hat{\omege}^{\eto})\ge t) - \PP( R(\hat{\omege}^{\ieo})\ge t) \\
\ge C - G^\ieo_{n,q,\sqrt{n}(t-\kappa_0^\ieo)} - G_{n,q}^\eto
\end{align*}


Herein, $C= 1- \exp\rbr{-\frac{n(\kappa_0^\eto - t)^2}{2\nbr{\nabla_\thete v_0(\omege_{\thete^\kl})\bM_1^\eto}^2}}$ if $\kappa_0^\ieo < t < \kappa_0^\eto$, $C= 0$ if $t > \kappa_0^\eto$, and $C = \frac{1}{2}$ if $t=\kappa_0^\eto$. $G_{n,q}^{\eto}$ and $G_{n,q,\sqrt{n}(t-\kappa_0^\ieo)}^{\ieo}$ are given in Proposition \ref{prop: BEforallRegret}. 
All the error terms $G_{n,q,\sqrt{n}(t-\kappa_0^\ieo)}^{\ieo}$, $G_{n,q}^\eto$ go to $0$ as $n\to \infty$. 

\end{theorem}


We discuss Theorem \ref{thm: finitecomparemis2}:

     \begin{enumerate}[leftmargin=*]
\item When $\kappa_0^\ieo<t < \kappa_0^\eto$, the benefit of IEO is outstanding, since $1$ is generally much larger than the statistical error terms $G^\ieo_{n,q,\sqrt{n}(t-\kappa_0^\ieo)} + G_{n,q}^\eto$ when $n$ is not too small. To provide intuition, $\PP(R(\hat{\omege}^{\eto})\ge t) \to 1$ as $n$ increases, since $R(\hat{\omege}^{\eto}) \xrightarrow{P} \kappa_0^{\eto} > t$. $\PP( R(\hat{\omege}^{\ieo})\ge t) \to 0$ as $n$ increases,  since $R(\hat{\omege}^{\ieo}) \xrightarrow{P} \kappa_0^{\ieo} < t$. This clearly shows that $\PP(R(\hat{\omege}^{\ieo})\ge t) \le \PP(R(\hat{\omege}^{\eto})\ge t)$ when $n$ is large. Therefore, even in a situation with very slight misspecification where $\mathbb G^{\eto}$ is first-order stochastically dominated by $\mathbb G^{\ieo}$ (Corollary \ref{cor: Ginmis}), we cannot claim the total regret of ETO is first-order stochastically dominated by that of IEO.

\item When $t > \kappa_0^\eto$, Theorem \ref{thm: finitecomparemis2} does not provide a strict distinction between IEO and ETO. However, we can still assert that the tail probability of the IEO regret is dominated by that of ETO, up to a finite-sample statistical error ($G^\ieo_{n,q,\sqrt{n}(t-\kappa_0^\ieo)} + G_{n,q}^\eto$). 
In other words, even in cases where IEO, if at all, performs worse than ETO, any potential performance degradation would be limited. 

\item When $t > \kappa_0^\eto$, the bound in Theorem \ref{thm: finitecomparemis2} and subsequent Theorem \ref{thm: finitecomparemis1} becomes trivial if we only consider $n \to \infty$. This is because when $t > \kappa_0^\eto$, both
$\PP(R(\hat{\omege}^{\eto})\ge t) \to 0 $ and $ \PP( R(\hat{\omege}^{\ieo})\ge t) \to 0$. Hence, while a standard asymptotic analysis cannot capture the subtle finite-sample tail difference in IEO and ETO, our Theorem \ref{thm: finitecomparemis2} and subsequent Theorem \ref{thm: finitecomparemis1} can as they hold for any $n\ge 1$.

\end{enumerate}  


Our next theorem shows that in a well-specified or slightly misspecified model, we can establish an upper bound for $\PP(R(\hat{\omege}^{\eto})\ge t) - \PP( R(\hat{\omege}^{\ieo})\ge t)$. Recall that another measurement of model specification besides $\delta$ is based on the eigenvalues of {$\bD^{\ieo}-\bD^{\eto}$}, which quantifies the difference in the second-order terms $\GG^\ieo$ and $\GG^\eto$, as shown in Theorem \ref{thm: Ginmis}. When $\tau_1\geq 0$, from the previous discussion, we have $\GG^\eto\preceq_\st \GG^\ieo$, which inherits the properties from the well-specified cases. In this case, ETO has some benefits in the $O(1/n)$ term, and such benefits can be formally stated in the following theorem.

\begin{theorem}[Upper Bound on $\cD$] \label{thm: finitecomparemis1}
Suppose Assumptions \ref{assumption: distributionmis}, \ref{assumption: Lipschitz Hessian of v0}, \ref{assumption: consistency all}, \ref{assumption: RCall}, and \ref{SCforh} hold. Suppose that $\tau_1 \ge 0$. For instance, this holds when ${B}_0 \le \tau_3/C_\mis$ by Corollary \ref{cor: Ginmis}. 
{Assume further that $H^\eto>0$.} In addition, let $\tau_6$ denote the largest eigenvalue of the matrix $\bM_1^{\eto}\nabla_{\thete\thete}\vale_0(\omege_{\thete^{\kl}}){\bM_1^{\eto}}$. Then we have the following results for any sample size $n\ge 1$.

Case 1: $t\le \kappa_0^{\ieo}:$
\begin{align*}
    \PP(R(\hat{\omege}^{\eto})\ge t) - \PP( R(\hat{\omege}^{\ieo})\ge t)=0
\end{align*}
Case 2: {$\delta>0$, $\tau_1>0$, and} $t> \kappa_0^{\ieo}+\frac{\tau_6+\tau_1}{\tau_1}\delta:$ for any $0< \varepsilon < \frac{\tau_1}{\tau_1+\tau_6} (t - \kappa_0^{\ieo}) - \delta$,
\begin{align*}
&\PP(R(\hat{\omege}^{\eto})\ge t) - \PP( R(\hat{\omege}^{\ieo})\ge t)\\
\le & -\PP(nt-n\kappa_0^\ieo \le \GG^{\ieo} \le \\&(1+\frac{\tau_1}{\tau_6})(nt- n\kappa_0^\ieo -n\delta-n\varepsilon))\\& + D_{n,q}^{\eto}+D_{n,q}^{\ieo} + C_{n,q}^{\eto} +  E_{n}^{\delta, \varepsilon}.
\end{align*}
Particularly when $\delta=0$, for any $t> \kappa_0^{\ieo}$,
\begin{align*}
&\PP(R(\hat{\omege}^{\eto})\ge t) - \PP( R(\hat{\omege}^{\ieo})\ge t)\\
\le & -\PP(nt-n\kappa_0^\ieo \le \GG^{\ieo} \le
(1+\frac{\tau_1}{\tau_6})(nt- n\kappa_0^\ieo))\\& + D_{n,q}^{\eto}+D_{n,q}^{\ieo}.
\end{align*}
Herein,
$$E_{n}^{\delta, \varepsilon} = \exp{\rbr{-\frac{n \varepsilon^2}{2\|\nabla_{\thete}\vale_0(\omege_{\thete^{\kl}}) {{\bM}_1^{\eto}}\|^2}}}$$
{is defined to be zero when the denominator is zero and $\varepsilon>0$, and otherwise} depends on the model misspecification. $D_{n,q}^{\eto}$, $D_{n,q}^{\ieo}$, $C_{n,q}^{\eto}$ are given by Propositions \ref{prop: BEforallRegret2} and \ref{prop: BEforall}. All the error terms $D_{n,q}^{\eto}$, $D_{n,q}^{\ieo}$, $C_{n,q}^{\eto}$, $E_{n}^{\delta, \varepsilon}$ go to $0$ as $n\to \infty$.

\end{theorem}

Note that Theorem \ref{thm: finitecomparemis2} provides a lower bound on $\cD=\PP(R(\hat{\omege}^{\eto})\ge t) - \PP( R(\hat{\omege}^{\ieo})\ge t)$ for all tail probabilities. In contrast, Theorem \ref{thm: finitecomparemis1} provides an upper bound on $\cD$, but, when $\delta>0$ and $\tau_1>0$, only for part of the regret tail probabilities (not covering $\kappa_0^{\ieo}< t \le \kappa_0^{\ieo}+\frac{\tau_6+\tau_1}{\tau_1}\delta$). We generally do not have a universal ordering in the intermediate region $\kappa_0^{\ieo} < t < \kappa_0^{\ieo}+\frac{\tau_6+\tau_1}{\tau_1}\delta$ when $\delta>0$ and $\tau_1>0$. However, when $\delta$ is sufficiently small, the intermediate region between Case 1 and Case 2 will be negligible. We discuss Theorem \ref{thm: finitecomparemis1} further:


 \begin{enumerate}[leftmargin=*]
\item When the model is well-specified, then $\delta = 0$, ${B}_0=0$ and $\tau_1\ge 0$. Then Theorem \ref{thm: finitecomparemis1} immediately applies to the well-specified case. In this case, Case 1 and Case 2 together cover all the tail probabilities, showing that any tail probability of ETO is less than that of IEO with a finite-sample statistical error ($D_{n,q}^{\eto}+D_{n,q}^{\ieo}$). This finite-sample estimation error will vanish as $n \to \infty$. 
To obtain an asymptotic result, letting $t= \frac{\tilde{t}}{n}$, we have
\begin{align*}
&\PP(R(\hat{\omege}^{\eto})\ge \frac{\tilde{t}}{n}) - \PP( R(\hat{\omege}^{\ieo})\ge \frac{\tilde{t}}{n})\\
\le & -\PP(\tilde{t} \le \GG^{\ieo} \le (1+\frac{\tau_1}{\tau_6})\tilde{t}) + D_{n,q}^{\eto}+D_{n,q}^{\ieo}.
\end{align*}
Note that the first term $\PP(\tilde{t} \le \GG^{\ieo} \le (1+\frac{\tau_1}{\tau_6})\tilde{t})$ is a constant independent of $n$, so taking $n\to \infty$, 
\begin{align*}
&\lim_{n\to \infty} \left(\PP(R(\hat{\omege}^{\eto})\ge \frac{\tilde{t}}{n}) - \PP( R(\hat{\omege}^{\ieo})\ge \frac{\tilde{t}}{n}) \right)\\
\le & -\PP(\tilde{t} \le \GG^{\ieo} \le (1+\frac{\tau_1}{\tau_6})\tilde{t}).
\end{align*}
This shows that asymptotically, $nR(\hat{\omege}^{\eto})$ is first-order stochastically dominated by $nR(\hat{\omege}^{\ieo})$, which leads to the asymptotic result in \citet{elmachtoub2023estimate}. 

\item When the model is misspecified, the double-benefit effect of IEO, as shown in Theorem \ref{thm: higherorder}, renders any potential advantage of ETO relatively minor. Moreover, such an advantage is only observed under a relatively restrictive setting. Theorem \ref{thm: finitecomparemis1} demonstrates that ETO exhibits a small advantage when model misspecification is zero or small, corresponding to a small or zero $\delta$ and $\tau_1 \geq 0$ stipulated by Corollary \ref{cor: Ginmis}. For $\delta>0$ and $\tau_1>0$, within a large tail region ($t > \kappa_0^{\ieo} + \frac{\tau_6 + \tau_1}{\tau_1} \delta$), the regret tail probability of ETO is always dominated by that of IEO, up to a finite-sample statistical error ($D_{n,q}^{\eto} + D_{n,q}^{\ieo} + C_{n,q}^{\eto}+  E_{n}^{\delta, \varepsilon}$). This suggests that ETO helps mitigate the occurrence of extremely large regrets compared to IEO.



\end{enumerate}

\section{Conclusions and Discussions} 
In this paper, we present a comprehensive theoretical analysis comparing the performance of IEO and ETO, focusing on how varying degrees of model misspecification influence their regrets across diverse scenarios. Our investigation employs several advanced techniques, including higher-order expansions of regret, the derivation of finite-sample regret bounds using recent Berry-Esseen results, and an in-depth characterization of regret tail behaviors in the finite-sample regime. 

Our results demonstrate that IEO consistently enjoys a ``universal double benefit'' in the two leading dominant terms of regret under general model misspecification. This fundamental advantage enables IEO to outperform ETO across a broad range of practical settings, providing statistical evidence for its superior empirical performance. On the other hand, when the underlying model is nearly well-specified, ETO could exhibit advantages thanks to its smaller estimation variability, and these advantages show up in the second-order term of regret.




\subsubsection*{Acknowledgements}
We gratefully acknowledge support from the National Science Foundation under grant CMMI-1763000, InnoHK initiative, the Government of the HKSAR, Laboratory for AI-Powered Financial Technologies, and Columbia SEAS Innovation Hub Award. The authors thank the anonymous reviewers for their constructive comments, which have greatly improved the quality of our paper.

\bibliographystyle{abbrvnat}
\bibliography{References}





\onecolumn
\setcounter{section}{0}
\renewcommand{\thesection}{\Alph{section}}

\aistatstitle{Dissecting the Impact of Model Misspecification in Data-Driven Optimization:
Supplementary Materials}


\section{Further Details and Proofs for Section \ref{sec: IEO generalization bounds}} \label{sec: proofs1}

\subsection{Examples of Parametric Distributions that Satisfy Assumption \ref{assumption: good parametrization}}\label{sec: good parametrization examples}

We show the generality of Assumption \ref{assumption: good parametrization} by giving concrete examples that satisfy the assumption. 
\begin{example}
    [Discrete Random Variables with Finite Support \citep{qi2021integrated} ] Suppose $\ran$ has finite discrete support, i.e., $\ran\in\cZ:=\cbr{\ran_1,\ran_2,...,{\ran_q}}$. The distribution of $\ran$ is characterized by a vector $\thete\in\Delta^{q-1}:=\{\thete\in\RR^q:\thete\geq 0,\sum_{k=1}^q\theta_k=1\}$. The total 
    \begin{align*}
        d_{\tv}(\PP_{\thete_1},\PP_{\thete_2})=\frac{1}{2}\sum_{k=1}^q|\theta_{1k}-\theta_{2k}|=\frac{1}{2}\|\thete_1-\thete_2\|_1\leq D_\Theta\|\thete_1-\thete_2\|.
    \end{align*}
\end{example}

\begin{example}
    [Multivariate Gaussian \citep{devroye2018total}] Consider the distribution family $\cP_{\Theta}:=\{N(\thete,\bSigma):\thete\in\Theta\}$ where $\bSigma$ is fixed{ and positive definite}. We have a closed form for the total variation distance:
    \begin{align*}
        d_{\tv}(N(\thete_1,\bSigma),N(\thete_2,\bSigma))&=\PP\left(N(0,1)\in\left[-\frac{\sqrt{(\thete_1-\thete_2)^\top\bSigma^{-1}(\thete_1-\thete_2)}}{2},\frac{\sqrt{(\thete_1-\thete_2)^\top\bSigma^{-1}(\thete_1-\thete_2)}}{2}\right]\right)\\
        &=\Phi\left(\frac{\|\thete_1-\thete_2\|_{\bSigma}}{2}\right)-\Phi\left(-\frac{\|\thete_1-\thete_2\|_{\bSigma}}{2}\right)
        \lesssim \|\thete_1-\thete_2\|_{\bSigma}\leq D_\Theta\|\thete_1-\thete_2\|.
    \end{align*}
\end{example}

\begin{example}
    [Mixture Gaussian \citep{doss2023optimal}] Let $\cG_{k,d}$ denote the collection of $k$-atomic distributions supported on a ball of radius $R$ is $d$ dimensions, i.e.,
    \begin{align*}
        \cG_{k,d}&:=\left\{\Gamma=\sum_{j=1}^kw_j\delta_{\mu_j}:\mu_j\in\RR^d, \|\mu_j\|_2\leq R,w_j\geq 0,\sum_{j=1}^kw_j=1\right\}.\\
        \cP_{k,d}&:=\{P_\Gamma:{\Gamma\in\cG_{k,d}}, P_\Gamma=\Gamma*N(0,\bI_d)\}.
    \end{align*}
    A parametrization that satisfies the identifiability assumption is provided by the moment tensors. The degree-$l$ moment tensor of the mixing distribution $\Gamma$ is the symmetric tensor
    \begin{align*}
        \cM_l(\Gamma):=\sum_{j=1}^kw_j\mu_j^{\otimes l}.
    \end{align*}
    It can be shown that any $k$-atomic distribution is uniquely determined by its first $2k-1$ moment tensors $\bm{\cM}_{2k-1}(\Gamma)=[\cM_1(\Gamma),...,\cM_{2k-1}(\Gamma)]$. Consequently, moment tensors provides a valid parametrization of the $k$-Gaussian-Mixture in the sense that $\bm{\cM}_{2k-1}(\Gamma)=\bm{\cM}_{2k-1}(\Gamma')\Leftrightarrow P_\Gamma=P_{\Gamma'}$. We have the following property:
    \begin{align*}
        d_{\tv}(P_\Gamma,P_\Gamma')\leq D_\Theta\max_{l\leq 2k-1}\|\cM_l(\Gamma)-\cM_l(\Gamma')\|_\textup{F}.
    \end{align*}
\end{example}


\begin{example}
    [Exponential Family \citep{busa2019optimal}] Let $\mu$ be a measure on $\RR^d$ and $h:\RR^d\to\RR_+,\bT:\RR^d\to\RR^k$ be measurable functions. 
    We define the logarithmic partition function ${\alpha_{\bT,h}:\RR^k\to\RR\cup\{+\infty\}}$ as $\alpha(\beeta)=\alpha_{\bT,h}(\beeta)=\ln(\int\exp(\beeta^\top\bT(\bx))h(\bx)d\mu(\bx))$. We also define the range of natural parameters $\cH_{\bT,h}={\{\beeta\in\RR^k:0<\int e^{\beeta^\top\bT(\bx)}h(\bx)d\mu(\bx)<\infty\}}$. The exponential family $\cE(\bT,h)$ with sufficient statistics $\bT$, carrier measure $h$ and natural parameters $\beeta$ is the family of distributions $\cE(\bT,h)=\{P_{\beeta}:\beeta\in\cH_{\bT,h}\}$ where the probability distribution $P_{\beeta}$ has density
    \begin{align*}
        p_{\beeta}(\bx)=\exp(\beeta^\top\bT(\bx)-\alpha(\beeta))h(\bx).
    \end{align*}

    It is known that the total variation of two distributions in the same exponential family has the following closed form. {Suppose $\Theta\subset\operatorname{int}(\cH_{\bT,h})$ is convex. For $\beeta,\beeta'\in\Theta$, there exists} $\bxi$ in the line segment of $[\beeta,\beeta']$ such that
    \begin{align*}
        d_{\tv}(P_{\beeta},P_{\beeta'})=\frac{1}{2}\underset{\bx\sim P_{\bxi}}{\EE}\left[\textup{sign}({p_{\beeta}(\bx)-p_{\beeta'}(\bx)})(\beeta-\beeta')^\top\left(\bT(\bx)-\underset{\by\sim P_{\bxi}}{\EE}[\bT(\by)]\right)\right].
    \end{align*}
    If we further assume the parameter space $\Theta\subset\cH_{\bT,h}$ is bounded and that $D_\Theta=\sup_{\beeta\in\Theta}\underset{\bx\sim P_{\beeta}}{\EE}(\|\bT(\bx)\|)<\infty$, then
        \begin{align*}
        d_{\tv}(P_{\beeta},P_{\beeta'})&=\frac{1}{2}\underset{\bx\sim P_{\bxi}}{\EE}\left[\textup{sign}({p_{\beeta}(\bx)-p_{\beeta'}(\bx)})(\beeta-\beeta')^\top\left(\bT(\bx)-\underset{\by\sim P_{\bxi}}{\EE}[\bT(\by)]\right)\right].\\
        &\leq\frac{1}{2}\underset{\bx\sim P_{\bxi}}{\EE}\left[\|\beeta-\beeta'\|\left\|\bT(\bx)-\underset{\by\sim P_{\bxi}}{\EE}[\bT(\by)]\right\|\right]\\
        &\leq\frac{1}{2}\underset{\bx\sim P_{\bxi}}{\EE}\left[\|\beeta-\beeta'\|\left(\left\|\bT(\bx)\right\|+\left\|\underset{\by\sim P_{\bxi}}{\EE}[\bT(\by)]\right\|\right)\right]\\&\leq D_\Theta\|\beeta-\beeta'\|.
    \end{align*}
    Concrete examples in the exponential family include: Univariate Gaussian, Multivariate Gaussian, Poisson, Centered Laplacian, Bernoulli, Binomial (with fixed number of trials), Multinomial (with fixed number of trials), negative binomial (with fixed number of failures) Rayleigh, Gamma, Beta, chi-squared, exponential, Dirichlet, geometric, etc.
\end{example}
More generally, if the density $P_\thete$ is uniformly smooth in the distribution family, which usually holds if $\Theta$ is bounded, then the distribution family satisfies Assumption \ref{assumption: good parametrization}, as presented in the following:
\begin{example}[Smooth Bounded Distribution Family]
    Suppose $\cZ$ is bounded{ and measurable, and the distributions have Lebesgue densities supported on $\cZ$} and $E_1:=\sup_{\ran\in\cZ}\|\ran\|$, and for all $\ran\in\cZ$, for all $P_\thete$, the density $p_\thete(\ran)$, as a function of $\thete$ is continuously differentiable with respect to $\thete${ on an open neighborhood of $\operatorname{conv}(\Theta)$}. Furthermore,$\sup_{\ran\in\cZ}{\sup_{\thete\in\operatorname{conv}(\Theta)}}\|\nabla_\thete p_\thete(\ran)\|=:E_2<\infty,$
    then for any $\ran\in\cZ$, $|p_{\thete_1}(\ran)-p_{\thete_2}(\ran)|\leq {E_2}\|\thete_1-\thete_2\|$.
This is because, by mean value theorem, there exists $\thete_3$ such that
\begin{align*}
    {|p_{\thete_1}(\ran)-p_{\thete_2}(\ran)|=\left|\nabla_\thete p_\thete(\ran)\big|_{\thete=\thete_3}(\thete_1-\thete_2)\right|}\leq\|\nabla_\thete {p_\thete(\ran)}|_{\thete=\thete_3}\|\|\thete_1-\thete_2\|\leq E_2\|\thete_1-\thete_2\|.
\end{align*}
Since $\cZ$ is bounded, then we can get an upper bound of the total variation
\begin{align*}
    2d_\tv(P_{\thete_1},P_{\thete_2})=\int_{\ran\in {\cZ}}|p_{\thete_1}(\ran)-p_{\thete_2}(\ran)|d\ran\leq\int_{\ran\in {\cZ}}E_2\|\thete_1-\thete_2\|d\ran\leq {E_2v_dE_1^d}\|\thete_1-\thete_2\|.
\end{align*}
In this case, it satisfies Assumption \ref{assumption: good parametrization} with $D_\Theta=\frac{1}{2}{E_2v_dE_1^d}$.
{Here $v_d$ is the volume of the Euclidean unit ball in $\RR^d$.}
\end{example}
Finally, any distribution family that satisfies Assumption \ref{assumption: good parametrization} can still preserve such property after truncation and normalization.
\begin{example}[Truncated Distribution Family]
    Suppose a distribution family {with densities $p_\thete$ }on $\cZ$ satisfies Assumption $\ref{assumption: good parametrization}$. The truncated distribution family $\tilde{p}_\thete$ defined on the restricted region $\tilde{\cZ}$ is given by $\tilde{p}_\thete(\ran)=1(\ran\in\tilde{\cZ})p_\thete(\ran)/{\int_{\ran\in\Tilde{\cZ}}p_\thete(\ran)\,d\ran}$. If we assume $\lambda:=\sup_{\thete\in\Theta}1/{\int_{\ran\in\Tilde{\cZ}}p_\thete(\ran)\,d\ran}{<\infty}$, then for any $\thete_1,\thete_2$, $d_\tv(\tilde{p}_{\thete_1},\tilde{p}_{\thete_2})\leq 2\lambda d_\tv({p}_{\thete_1},{p}_{\thete_2})\leq {2D_\Theta\lambda}\|\thete_1-\thete_2\|$. To see this, define $\lambda_1=1/{\int_{\ran\in\Tilde{\cZ}}p_{\thete_1}(\ran)\,d\ran}$ and $\lambda_2=1/{\int_{\ran\in\Tilde{\cZ}}p_{\thete_2}(\ran)\,d\ran}$ and $\lambda_1\leq\lambda_2\leq\lambda$ without loss of generality.We define $\cZ_1:=\{\ran\in\Tilde{\cZ}:{\lambda_1p_{\thete_1}(\ran)}>{\lambda_2p_{\thete_2}(\ran)}\}$ and ${\cZ_2}:=\{\ran\in\Tilde{\cZ}:{\lambda_1p_{\thete_1}(\ran)}\leq{\lambda_2p_{\thete_2}(\ran)}\}$. The total variation satisfies
    \begin{align*}
        2d_\tv(\tilde{p}_{\thete_1},\tilde{p}_{\thete_2})&=\int_{\ran\in\tilde{\cZ}}|\lambda_1p_{\thete_1}(\ran)-\lambda_2p_{\thete_2}(\ran)|d\ran\\&=\int_{\ran\in\cZ_1}(\lambda_1p_{\thete_1}(\ran)-\lambda_2p_{\thete_2}(\ran))d\ran+\int_{\ran\in\cZ_2}(\lambda_2p_{\thete_2}(\ran)-\lambda_1p_{\thete_1}(\ran))d\ran\\
        &=2\int_{\ran\in\cZ_1}(\lambda_1p_{\thete_1}(\ran)-\lambda_2p_{\thete_2}(\ran))d\ran\leq2\int_{\ran\in\cZ_1}(\lambda_2p_{\thete_1}(\ran)-\lambda_2p_{\thete_2}(\ran))d\ran\\
        &=2\int_{\ran\in\cZ_1}|\lambda_2p_{\thete_1}(\ran)-\lambda_2p_{\thete_2}(\ran)|d\ran\leq 2\int_{\ran\in\cZ}|\lambda_2p_{\thete_1}(\ran)-\lambda_2p_{\thete_2}(\ran)|d\ran\\
        &=4\lambda_2d_\tv(p_{\thete_1},p_{\thete_2})\leq 4\lambda d_\tv(p_{\thete_1},p_{\thete_2}).
    \end{align*}
    In most cases, the distribution family may be truncated to the concentrated region so that $\lambda$ will be slightly larger than $1$.
\end{example}

\subsection{Proof of Lemma \ref{lemma: uniform convergence} and Theorem \ref{thm: IEO risk bound}}\label{subsec: proof of theorem of ieo risk bound}
Since the expected cost function $\vale(\omege,Q)$ (resp. $\vale(\omege,\thete)$) is strongly convex for any $Q\in\cP$ (resp. $\thete\in\Theta$), we have the following first order property, as presented in Lemma \ref{lemma: strong cvx first order}.
\begin{lemma}[Theorem 5.24 in \cite{beck2017first}]
\label{lemma: strong cvx first order} Under Assumption \ref{assumption: convexity and strong convexity of cost function}{ and the convexity of $\Omega$}, for any $Q\in\cP$, for any $\omege\in\Omega$ and for any $\thete\in\Theta$
    \begin{align*}
        v(\omege,Q)-v(\omege_{Q},Q)&\geq {\frac{\rho_c}{2}\|\omege-\omege_{Q}\|^2}.\\
    v(\omege,\thete)-v(\omege_\thete,\thete)&\geq {\frac{\rho_c}{2}\|\omege-\omege_\thete\|^2}.
\end{align*}
\end{lemma}
{Indeed, for either objective $f$ with minimizer $\omege_0$, strong convexity gives $f(\omege_0)\leq f((1-t)\omege_0+t\omege)\leq(1-t)f(\omege_0)+tf(\omege)-\rho_ct(1-t)\|\omege-\omege_0\|^2/2$. Divide by $t>0$ and let $t\downarrow0$.}
To establish the generalization bound for IEO, we rely on both single-variate and multi-variate Rademacher complexity. Given a distribution family $\cbr{P_\thete:\thete\in\Theta}$, we can apply generalization bounds that directly use the Rademacher complexity of the function class $\cost\circ\Theta$. Given a sample $\cbr{\ran_i}_{i=1}^n$, the \textit{empirical Rademacher complexity} $\radhat_n^\ieo(\Theta)$ of the function class $\cost\circ\Theta$ is defined by
\begin{align*}
\hat{\rad}_n^\ieo(\Theta)&:=\EE_\sigma\left[\sup_{\thete\in\Theta}\frac{1}{n}\sum_{i=1}^n\sigma_ic(\omege_\thete,\ran_i)\right],
\end{align*}
where $\sigma_i$ are independent Rademacher random variables, i.e. $\PP(\sigma_i=1)=\PP(\sigma_i=-1)=\frac{1}{2}$ for all $i\in[n]$. The \textit{expected Rademacher complexity} $\rad_n^\ieo(\Theta)$ is then defined as the expectation of $\radhat_n^\ieo(\Theta)$ with respect to the i.i.d. sample $\cbr{\ran_i}_{i=1}^n$ drawn from the distribution $P$:
\begin{align*}
    {\rad}_n^\ieo(\Theta)&:=\EE_{\ran\sim P}\left[\hat{\rad}_n^\ieo(\Theta)\right].
\end{align*}
By directly using ${\rad}_n^\ieo(\Theta)$, we have the following theorem is an adaptation of the classical generalization bounds based on Rademacher complexity due to \cite{mohri2018foundations} to our setting.


\begin{lemma}[Theorem 3.3 in \cite{mohri2018foundations}]\label{lemma: classical generalization}
    Under Assumption \ref{assumption: boundedness of cost function}, for any ${0<\tilde{\delta}<1}$, with probability at least $1-\tilde{\delta}$ over an i.i.d. sample of $\{\ran_i\}_{i=1}^n$ drawn from $P$, the following holds for all $\thete\in\Theta$:
    \begin{align*}
v_0(\omege_\thete)\leq\hat{v}_0(\omege_\thete)+2\rad_n^\ieo(\Theta)+B_c\sqrt{\frac{\log(1/\tilde{\delta})}{2n}}.
    \end{align*}
\end{lemma}
Equipped with Lemma \ref{lemma: classical generalization}, which holds for all $\thete\in\Theta$, we have the following guarantee on the excess risk of the empirical minimizer of IEO $\hat{\omege}^\ieo$. Corollary \ref{cor: risk bounds} below is the combination of Lemma \ref{lemma: classical generalization} and the Hoeffding inequality.

\begin{corollary}
    \label{cor: risk bounds}
    Under Assumption \ref{assumption: boundedness of cost function}, for any ${0<\tilde{\delta}<1}$, with probability at least $1-\tilde{\delta}$ over an i.i.d. sample of $\{\ran_i\}_{i=1}^n$ drawn from $P$, the decision $\hat{\omege}^{\ieo}$ returned by the IEO satisfies:
    \begin{align*}
        R(\hat{\omege}^\ieo)\leq R(\omege_{\thete^*})+2\rad_n^\ieo(\Theta)+2B_c\sqrt{\frac{\log(2/\tilde{\delta})}{2n}}.
    \end{align*}
\end{corollary}
\proof[Proof of Corollary \ref{cor: risk bounds}]
By the definition of the regret of $\hat{\omege}^\ieo$,
\begin{align*}
    R(\hat{\omege}^\ieo){-R(\omege_{\thete^*})}&=v_0(\hat{\omege}^\ieo)-v_0(\omege_{\thete^*})\\&=v_0(\hat{\omege}^\ieo)-\hat{v}_0(\hat{\omege}^\ieo)+\hat{v}_0(\hat{\omege}^\ieo)-\hat{v_0}(\omege_{\thete^*})+\hat{v_0}(\omege_{\thete^*})-v_0(\omege_{\thete^*})\\
    &\leq v_0(\hat{\omege}^\ieo)-\hat{v}_0(\hat{\omege}^\ieo)+\hat{v_0}(\omege_{\thete^*})-v_0(\omege_{\thete^*}).
\end{align*}
The inequality follows from the fact that $\hat{\thete}^{\ieo}$ is the empirical minimizer of $\hat{v}_0(\omege_\thete)$. From Lemma \ref{lemma: classical generalization} we know that
\begin{align*}
v_0(\hat{\omege}^\ieo)\leq\hat{v}_0(\hat{\omege}^\ieo)+2\rad_n^\ieo(\Theta)+B_c\sqrt{\frac{\log(2/\tilde{\delta})}{2n}}
\end{align*}
with probability at least $1-\tilde{\delta}/2$. From Hoeffding's inequality, we have that
\begin{align*}
\hat{v}_0(\omege_{\thete^*})-v_0(\omege_{\thete^*})\leq B_c\sqrt{\frac{\log(2/\tilde{\delta})}{2n}}
\end{align*}
with probability at least $1-\tilde{\delta}/2$. Thus, with probability at least $1-\tilde{\delta}$ we can combine the three previous inequalities and obtain the desired result.
\endproof

Next, we introduce the multivariate Rademacher complexity as an extension of the regular Rademacher complexity to a class of vector-valued functions. Given a fixed sample $\{(\omege_\thete,\ran_i)\}_{i=1}^n$, we define the \textit{empirical multivariate Rademacher complexity} of $\Theta$ as
\begin{align*}
    \hat{\rad}^n_{\omege}(\Theta)&:=\EE_\sigma\left[\sup_{\thete\in\Theta}\frac{1}{n}\sum_{i=1}^n\sum_{j=1}^p\sigma_{ij}^\top\omege_{\thete j}\right]=\EE_\sigma\left[\sup_{\thete\in\Theta}\frac{1}{n}\sum_{i=1}^n\bsigma_i^\top\omege_\thete\right].
\end{align*}
In this case, $\hat{\rad}^n_{\omege}(\Theta)$ is deterministic and it always holds that the \textit{expected multivariate Rademacher complexity} ${\rad}^n_{\omege}(\Theta):=\underset{\ran\sim P}{\EE}\hat{\rad}^n_{\omege}(\Theta)\equiv\hat{\rad}^n_{\omege}(\Theta)$. A famous result of the multivariate Rademacher complexity is the \textit{vector contraction inequalities}.  Let $\Phi_i:\RR^p\to\RR$ for $i\in[n]$ be a collection of $L$-Lipschitz functions with respect to the give norm $\|\cdot\|$ defined on $\RR^p$:
\begin{align*}
    |\Phi_{i}(\bu)-\Phi_{i}(\bv)|\leq L\|\bu-\bv\| \ \ \text{ for all ${\bu,\bv\in \RR^p}$}.
\end{align*}
A vector contraction inequality \citep{maurer2016vector} takes the form:
\begin{align*}
    \EE_{\sigma}\left[\sup_{\thete\in\Theta}\frac{1}{n}\sum_{i=1}^n\sigma_i\Phi_{i}(\omege_\thete)\right]\leq CL\cdot\EE_\sigma\left[\sup_{\thete\in\Theta}\frac{1}{n}\sum_{i=1}^n\bsigma_i^\top\omege_\thete\right]=CL\cdot\hat{\rad}^n_{\omege}(\Theta),
\end{align*}
where $C$ is a constant. In our setting, $\Phi_i(\cdot)=\cost(\cdot,\ran_i)$ so $\Phi_i(\cdot)$ is $L$-Lipschitz for all $i\in[n]$ by Assumption \ref{assumption: Lipschitzness of cost function}. If $\|\cdot\|$ is $\ell_2$ norm, we can apply an elegant vector contraction inequality due to \cite{maurer2016vector}, which exactly takes the form $C=\sqrt{2}$. We have the following relations between $\rad_n^\ieo(\Theta)$ and $\rad^n_{\omege}(\Theta)$.
\begin{lemma}
[\bf{Corollary 4 in \cite{maurer2016vector}}] Under Assumption \ref{assumption: Lipschitzness of cost function}, $ \rad_n^\ieo(\Theta)\leq \sqrt{2}L_c\rad^n_{\omege}(\Theta)$.
\end{lemma}

We are now going to prove a Lipschitz property of the optimization oracle $\omege_\thete$ with respect to the total variation distances of two distributions. We will prove a more general result. Let $\omege_{Q}$ denote the optimization oracle, i.e., $\omege_{Q}:=\argmin_{\omege\in\Omega}\underset{\ran\sim Q}{\EE}[c(\omege,\ran)]$. We show that $\omege_\thete$ is ``Lipschitz'' with respect to $\thete$ in some sense.
\begin{lemma}\label{lemma: TV Lipschitz continuity}
    Under Assumption \ref{assumption: convexity and strong convexity of cost function} and \ref{assumption: Lipschitzness of cost function}{, with $\Omega$ convex}, for any two distributions $Q_1,Q_2\in\cP$, 
    \begin{enumerate}[leftmargin=*]
        \item If they have density $f_{Q_1}$ and $f_{Q_2}${ with respect to a common dominating measure $\mu$ (which can always be chosen as $Q_1+Q_2$)}, then $\|\omege_{{Q_1}}-\omege_{Q_2}\|\leq\frac{2L_c}{\rho_c}d_{\tv}(Q_1,Q_2)$.
        \item If they have the same finite support $\{\ran_k\}_{k=1}^K$, with probability mass function $\bq^1, \bq^2\in \Delta^{K-1}$, then $\|\omege_{{Q_1}}-\omege_{Q_2}\|\leq\frac{2L_c}{\rho_c}d_{\tv}(Q_1,Q_2)$.
        \item If the parametric family $\cbr{P_\thete:\thete\in\Theta}$ further satisfies Assumption \ref{assumption: good parametrization}, then $\|\omege_{\thete_1}-\omege_{\thete_2}\|\leq \frac{2L_cD_\Theta}{\rho_c}\|\thete_1-\thete_2\|$. 
    \end{enumerate}

\end{lemma}
\proof[Proof of Lemma \ref{lemma: TV Lipschitz continuity}]
    By the strong convexity of Assumption \ref{assumption: convexity and strong convexity of cost function}, for $Q_1,Q_2\in\cP$, 
    \begin{align*}
         v(\omege_{Q_2},Q_1)-v(\omege_{Q_1},Q_1)&\geq \frac{\rho_c}{2}\|{\omege_{Q_2}}-\omege_{Q_1}\|^2\\
         v(\omege_{Q_1},Q_2)-v(\omege_{Q_2},Q_2)&\geq \frac{\rho_c}{2}\|{\omege_{Q_1}}-\omege_{Q_2}\|^2.
    \end{align*}
    In the first scenario,
    \begin{align*}
        \rho_c\|\omege_{Q_2}-\omege_{Q_1}\|^2&\leq v(\omege_{Q_2},Q_1)-v(\omege_{Q_1},Q_1)+v(\omege_{Q_1},Q_2)-v(\omege_{Q_2},Q_2)\\
        &=\int c(\omege_{Q_2},\ran)d Q_1-c(\omege_{Q_1},\ran)d Q_1+c(\omege_{Q_1},\ran)d Q_2-c(\omege_{Q_2},\ran)d Q_2\\
        &= \int \rbr{c(\omege_{Q_2},\ran)-c(\omege_{Q_1},\ran)}(d Q_1-d Q_2)\\
        &=\int_{\ran\in\cZ} (c(\omege_{Q_2},\ran)-c(\omege_{{Q_1}},\ran))(f_{Q_1}(\ran)-f_{Q_2}(\ran)){d\mu(\ran)}\\
        &\leq \int_{\ran\in\cZ} |c(\omege_{Q_2},\ran)-c(\omege_{Q_1},\ran)||f_{Q_1}(\ran)-f_{Q_2}(\ran)|{d\mu(\ran)}\\
        &\leq {L_c}\|\omege_{Q_2}-\omege_{Q_1}\|\int_{\ran\in\cZ}|f_{Q_1}(\ran)-f_{Q_2}(\ran)|{d\mu(\ran)}\\
        &=2{L_c}\|\omege_{Q_2}-\omege_{Q_1}\|d_{\tv}({Q_1},{Q_2}).
    \end{align*}
    In the second scenario,
    \begin{align*}
        \rho_c\|\omege_{Q_2}-\omege_{Q_1}\|^2&\leq v(\omege_{Q_2},Q_1)-v(\omege_{Q_1},Q_1)+v(\omege_{Q_1},Q_2)-v(\omege_{Q_2},Q_2)\\
        &=\sum_{k=1}^Kq^1_kc(\omege_{Q_2},\ran_k)-\sum_{k=1}^Kq^1_kc(\omege_{Q_1},\ran_k)+\sum_{k=1}^Kq^2_kc(\omege_{Q_1},\ran_k)-\sum_{k=1}^Kq^2_kc(\omege_{Q_2},\ran_k)\\
        &=\sum_{k=1}^K(q^1_k-q^2_k)\left(c(\omege_{Q_2},\ran_k)-c(\omege_{Q_1},\ran_k)\right)\\
         &\leq\sum_{k=1}^K|q^1_k-q^2_k|{\left|c(\omege_{Q_2},\ran_k)-c(\omege_{Q_1},\ran_k)\right|}\\
         &\leq\sum_{k=1}^K|q^1_k-q^2_k| L_c\|\omege_{Q_2}-\omege_{Q_1}\|\\
         &=2L_c\|\omege_{Q_2}-\omege_{Q_1}\|d_{\tv}({Q_1},{Q_2}).
    \end{align*}
In particular, the result holds for a distribution family $\cP_{\Theta}$ that satisfies either scenario in Lemma \ref{lemma: TV Lipschitz continuity}: for any $\thete_1,\thete_2\in\Theta, \|\omege_{\thete_1}-\omege_{\thete_2}\|\leq\frac{2L_c}{\rho_c}d_{\tv}(P_{\thete_1},P_{\thete_2})$. If the parametric family further satisfies Assumption \ref{assumption: good parametrization}, then for any $\thete_1,\thete_2\in\Theta, \|\omege_{\thete_1}-\omege_{\thete_2}\|\leq\frac{2L_c}{\rho_c}d_{\tv}(P_{\thete_1},P_{\thete_2})\leq \frac{2L_cD_\Theta}{\rho_c}\|\thete_1-\thete_2\|$.
\endproof

In the remainder of this section, we will leverage the \textit{covering number} argument to bound $\rad^n_{\omege}(\Theta)$. We first review some basic ingredients of covering number.
\begin{definition}[Covering]\label{def: packing and covering}
    Consider the metric space $(\RR^q,\|\cdot\|)$ and $\Theta\subset\RR^q$. 
    \begin{enumerate}[leftmargin=*]
        \item We say $\{\thete_1,...,\thete_N\}\subset\RR^q$ is an \textit{$\varepsilon$-covering} of ${\Theta}$ if for all $\thete\in\Theta$, there exists $i\in[N]$ such that $\|\thete_i-\thete\|<\varepsilon$. 
        \item The \textit{$\varepsilon$-covering number} of ${\Theta}$ is
$ \cN(\varepsilon,\Theta,\|\cdot\|):=\inf\{N\in\ZZ^+: \exists\ \text{an $\varepsilon$-covering}\ \thete_1,...,\thete_N \ \text{of}\  {\Theta}\}.$
    \end{enumerate}
\end{definition}
\begin{lemma}\label{lemma: covering number of Theta}When $\Theta$ is bounded, we denote $E_{\Theta}:=\sup_{\thete\in\Theta}\|\thete\|$. For $\varepsilon>0$, the covering number of $\Theta$ satisfies
    \begin{align*}
        \cN(\varepsilon,\Theta,\|\cdot\|)\leq {\left(1+\frac{2E_\Theta}{\varepsilon}\right)^q}.
    \end{align*}
\end{lemma}
\proof[Proof of Lemma \ref{lemma: covering number of Theta}]
By Lemma 6.27 in \cite{mohri2018foundations}, for any $\varepsilon>0$    
\begin{align*}
        \left(\frac{1}{\varepsilon}\right)^q\leq\cN(\varepsilon,B(0,1),\|\cdot\|)\leq \left(1+\frac{2}{\varepsilon}\right)^q.
    \end{align*}
    For general bounded set $\Theta$, it satisfies
    \begin{align*}
        {\cN(\varepsilon,\Theta,\|\cdot\|)
        \leq\cN(\varepsilon,B(0,E_\Theta),\|\cdot\|)
        =\cN(\varepsilon/E_\Theta,B(0,1),\|\cdot\|)
        \leq\left(1+\frac{2E_\Theta}{\varepsilon}\right)^q.}
    \end{align*}
    {If $E_\Theta=0$, the bound is immediate.}
\endproof

We are now ready to provide an explicit upper bound of $\rad^n_\omege(\Theta)$:
\begin{lemma}\label{lemma: bound on multivariate complexity}
    There exists a universal constant $C_\textup{abs}$ such that the multivariate Rademacher complexity $\rad^n_\omege(\Theta)$ satisfies
    \begin{align*}
        \rad^n_\omege(\Theta)\leq \frac{2L_cC_\textup{abs}D_\Theta E_{\Theta}}{\rho_c}\sqrt{\frac{q}{n}}.
    \end{align*}
\end{lemma}
\proof[Proof of Lemma \ref{lemma: bound on multivariate complexity}]
For simplicity, we denote $L':=2L_cD_\Theta/\rho_c$, $X_\thete:=\frac{1}{\sqrt{n}}\sum_{i=1}^n\bsigma_i^\top\omege_\thete$. {The case $L' E_\Theta=0$ is immediate; assume otherwise.} It is not hard to see that since for any $\thete_1,\thete_2\in\Theta$, 
\begin{align*}
    X_{\thete_1}-X_{\thete_2}=\sum_{i=1}^n\frac{1}{\sqrt{n}}\bsigma_i^\top(\omege_{\thete_1}-\omege_{\thete_2})=\sum_{i=1}^n\sum_{j=1}^p\sigma_{ij}\frac{1}{\sqrt{n}}(\omege_{\thete_1j}-\omege_{\thete_2j}).
\end{align*}
By Lemma 3.12 in \cite{sen2018gentle}, $X_{\thete_1}-X_{\thete_2}$ is $\|\omege_{\thete_1}-\omege_{\thete_2}\|^2$ sub-Gaussian, and thus $L'^2\|\thete_1-\thete_2\|^2$ sub-Gaussian. In conclusion, $\cbr{X_\thete}_{\thete\in\Theta}$ is a sub-Gaussian process. By Theorem 4.3 in \cite{sen2018gentle}, for any fixed $\thete_0\in\Theta$,
\begin{align*}
    \EE\sup_{\thete\in\Theta}X_\thete&=\EE\sup_{\thete\in\Theta}(X_\thete-X_{\thete_0})\leq \EE\sup_{\thete\in\Theta}|X_\thete-X_{\thete_0}|\lesssim\int_{0}^\infty\sqrt{\log\cN(\varepsilon,\Theta,L'\|\cdot\|)}d\varepsilon\\
    &={\int_0^{2L'E_\Theta}\sqrt{\log\cN(\varepsilon,\Theta,L'\|\cdot\|)}d\varepsilon}\\
    &={L'\int_0^{2E_\Theta}\sqrt{\log\cN(\varepsilon,\Theta,\|\cdot\|)}d\varepsilon}\\
    &\leq {L'\sqrt q\int_0^{2E_\Theta}\sqrt{\log(1+2E_\Theta/\varepsilon)}d\varepsilon}\\
    &={2L'E_\Theta\sqrt q\int_0^1\sqrt{\log(1+1/u)}du}\\
    &\leq C_\textup{abs}L'\sqrt{q}{E_{\Theta}}.
\end{align*}

The multivariate Rademacher complexity thus satisfies
\begin{align*}
    \rad^n_\omege(\Theta)=\frac{1}{\sqrt{n}}\EE\sup_{\thete\in\Theta}X_\thete\leq C_\textup{abs}L'{E_{\Theta}}\sqrt{\frac{q}{n}}=\frac{2L_cC_\textup{abs}D_{\Theta}{E_{\Theta}}}{\rho_c}\sqrt{\frac{q}{n}}.
\end{align*}
\endproof
Finally, we can prove Lemma \ref{lemma: uniform convergence} and Theorem \ref{thm: IEO risk bound}:
\proof[Proof of Lemma \ref{lemma: uniform convergence}]
By previous discussion, we have
    \begin{align*}
v_0(\omege_\thete)&\leq\hat{v}_0(\omege_\thete)+2\rad_n^\ieo(\Theta)+B_c\sqrt{\frac{\log(1/\tilde{\delta})}{2n}}\\
&\leq \hat{v}_0(\omege_\thete)+\frac{4\sqrt{2}L_c^2{C_\textup{abs}}D_{\Theta}{E_{\Theta}}}{\rho_c}\sqrt{\frac{q}{n}}+B_c\sqrt{\frac{\log(1/\tilde{\delta})}{2n}}. 
    \end{align*}
\endproof
\proof[Proof of Theorem \ref{thm: IEO risk bound}]
By Corollary \ref{cor: risk bounds}, we have
    \begin{align*}
        R(\hat{\omege}^\ieo)&\leq R(\omege_{\thete^*})+2\rad_n^\ieo(\Theta)+2B_c\sqrt{\frac{\log(2/\tilde{\delta})}{2n}}\\
        &\leq R(\omege_{\thete^*})+2\sqrt{2}L_c\rad^n_{\omege}(\Theta)+2B_c\sqrt{\frac{\log(2/\tilde{\delta})}{2n}}\\
        &\leq R(\omege_{\thete^*})+\frac{4\sqrt{2}L_c^2C_\textup{abs}D_{\Theta}{E_{\Theta}}}{\rho_c}\sqrt{\frac{q}{n}}+2B_c\sqrt{\frac{\log(2/\tilde{\delta})}{2n}}.
    \end{align*}
\endproof


\section{Further Details and Proofs in Section \ref{sec: comparison ETO and IEO}}\label{sec: further details in comparison}

\subsection{Assumptions and Supporting Theorems}

{Throughout this section, $\Theta$ is convex, criteria are finite and measurable, population optima lie in $\Theta$, and empirical optimizers are measurable and attained almost surely. In the regularity assumptions, parameter derivatives are continuous and commute with expectation; matrix norms are operator norms.}

We first list out standard assumptions that lead to the consistency of the ETO and IEO, which are direct consequences of asymptotic statistical theory (e.g., Theorem 5.7 in \citet{van2000asymptotic}.)

\begin{subtheorem}{assumption}\label{assumption: consistency all}
\begin{assumption}[Consistency conditions for ETO]\label{assumption: ETOconsistency}
Suppose that:

\begin{enumerate}
\item $\sup_{\thete\in \Theta} |\frac{1}{n}\sum_{i=1}^n \log p_\thete (\ran_i)-\mathbb{E}_{P} [ \log p_\thete(\ran)] |\xrightarrow{P} 0$.
\item $\forall\varepsilon> 0$, $\sup_{\thete\in \Theta: \|\thete-\thete^{\kl}\|\ge \varepsilon} \mathbb{E}_{P} [ \log p_\thete(\ran)] < \mathbb{E}_{P} [ \log p_{\thete^{\kl}}(\ran)]$.
\end{enumerate}
\end{assumption}

\begin{assumption}[Consistency conditions for IEO]\label{assumption: IEOconsistency}
Suppose that:

\begin{enumerate}
\item $\sup_{\thete\in \Theta} |\hat{\vale}_0(\omege_\thete)-\vale_0(\omege_\thete)|\xrightarrow{P} 0$. 
\item $\forall\varepsilon> 0$,
$\inf_{\thete\in \Theta: \|\thete-\thete^*\|\ge \varepsilon} \vale_0(\omege_\thete) > \vale_0(\omege_{\thete^*})$. 
\end{enumerate}
\end{assumption}
\end{subtheorem}

In each of Assumptions \ref{assumption: ETOconsistency} and \ref{assumption: IEOconsistency}, the first part is a uniform law of large numbers that are satisfied via Glivenko-Cantelli conditions. The second part ensure the uniqueness of $\thete^{\kl}$ or $\thete^*$. 

We also list out standard conditions and asymptotic normality guarantees for IEO and ETO which follow directly from established results in asymptotic statistical theory (e.g. \citet{shao2022berry}.)

\begin{subtheorem}{assumption}\label{assumption: RCall}
    \begin{assumption}[Regularity conditions for ETO]\label{assumption: RCforETO}
        The function $\log p_{\thete}(\cdot)$ is twice differentiable with respect to $\thete$ and there exist constant $\mu^\eto>0, c_1^\eto>0, c_2^\eto>0$ and two nonnegative functions $K_1^\eto, K_2^\eto:\cZ\to\RR$ with $\|K_1^\eto(\ran)\|_9\leq c_1^\eto$ and $\|K_2^\eto(\ran)\|_4\leq c_2^\eto$ such that for any $\thete\in\Theta$,
        \begin{align*}
            {\EE_P(\log p_{\thete^{\kl}}(\ran))-\EE_P(\log p_\thete(\ran))}&\geq \mu^\eto\|\thete-\thete^\kl\|^2,\\
            |\log p_{\thete}(\ran)-\log P^{\kl}(\ran)|&\leq K_1^\eto(\ran)\|\thete-\thete^\kl\|, \quad \forall \ran\in\cZ,\\
            \|\nabla_{\thete\thete}\log p_{\thete}(\ran)-{\nabla_{\thete\thete}\log p_{\thete^{\kl}}(\ran)}\|&\leq K_2^\eto(\ran)\|\thete-\thete^\kl\|, \quad \forall \ran\in\cZ.
        \end{align*}
        Moreover, there exists a constant $c_3^\eto\geq 0$ and a nonnegative function $K_3^\eto:\cZ\to\RR$ such that for any $\ran\in\cZ$:
        \begin{align*}
            {\|\nabla_{\thete\thete}\log p_{\thete^{\kl}}(\ran)\|\leq K_3^\eto(\ran)}\ \textup{and}\ \|K_3^\eto(\ran)\|_4\leq c_3^\eto.
        \end{align*}
        Let $\Sigma^\eto=\EE[\nabla_\thete \log p_{\thete^{\kl}}(\ran)^\top\nabla_\thete \log p_{\thete^{\kl}}(\ran)]$ and $V^\eto={-}\nabla_{\thete\thete}\EE[\log p_{\thete^{\kl}}(\ran)]$. Assume that there exist constants $\lambda_1^\eto>0$ and $\lambda_2^\eto>0$ such that $\lambda_{\min}(\Sigma^\eto)\geq \lambda_1^\eto$ and $\lambda_{\min}(V^\eto)\geq \lambda_2^\eto$. Moreover, assume that there exists a constant $c_4^\eto>0$ such that
        \begin{align*}
            \|\nabla_\thete \log p_{\thete^{\kl}}(\ran)\|_4\leq c_4^\eto\sqrt{q}.
        \end{align*}
    \end{assumption}

    \begin{assumption}[Regularity conditions for IEO]\label{assumption: RCforIEO}
        The function $\cost(\omege_\thete,\cdot)$ is twice differentiable with respect to $\thete$ and there exist constant $\mu^\ieo>0, c_1^\ieo>0, c_2^\ieo>0$ and two nonnegative functions $K_1^\ieo, K_2^\ieo:\cZ\to\RR$ with $\|K_1^\ieo(\ran)\|_9\leq c_1^\ieo$ and $\|K_2^\ieo(\ran)\|_4\leq c_2^\ieo$ such that for any $\thete\in\Theta$,
        \begin{align*}
            v_0(\omege_\thete)-v_0(\omege_{\thete^*})&\geq \mu^\ieo\|\thete-\thete^*\|^2,\\
            |\cost(\omege_\thete,\ran)-\cost({\omege_{\thete^*}},\ran)|&\leq K_1^\ieo(\ran)\|\thete-\thete^*\|, \quad \forall \ran\in\cZ,\\
            \|\nabla_{\thete\thete}\cost(\omege_\thete,\ran)-\nabla_{\thete\thete}\cost(\omege_{\thete^*},\ran)\|&\leq K_2^\ieo(\ran)\|\thete-\thete^*\|, \quad \forall \ran\in\cZ.
        \end{align*}
        Moreover, there exists a constant $c_3^\ieo\geq 0$ and a nonnegative function $K_3^\ieo:\cZ\to\RR$ such that for any $\ran\in\cZ$:
        \begin{align*}
            {\|\nabla_{\thete\thete}c(\omege_{\thete^*},\ran)\|\leq K_3^\ieo(\ran)}\ \textup{and}\ \|K_3^\ieo(\ran)\|_4\leq c_3^\ieo.
        \end{align*}
        Let $\Sigma^\ieo=\EE[\nabla_\thete c(\omege_{\thete^*},\ran)^\top\nabla_\thete c(\omege_{\thete^*},\ran)]$ and $V^\ieo=\nabla_{\thete\thete}\EE[c(\omege_{\thete^*},\ran)]$. Assume that there exist constants $\lambda_1^\ieo>0$ and $\lambda_2^\ieo>0$ such that $\lambda_{\min}(\Sigma^\ieo)\geq \lambda_1^\ieo$ and $\lambda_{\min}(V^\ieo)\geq \lambda_2^\ieo$. Moreover, assume that there exists a constant $c_4^\ieo>0$ such that
        \begin{align*}
            \|\nabla_\thete c(\omege_{\thete^*},\ran)\|_4\leq c_4^\ieo\sqrt{q}.
        \end{align*}
    \end{assumption}
\end{subtheorem}

We give some remarks on the twice differentiability in the assumptions, e.g. in Assumption \ref{assumption: RCforIEO}. (1) The twice differentiability of $c(\omege_{\thete},\ran)$ holds for many applications. However, even if sometimes $\nabla_{\thete\thete}c(\omege_{\thete},\ran)$ does not exist, Berry-Esseen bounds can still be established under some Lipschitz continuous gradient assumptions (Theorem 3.2 in \cite{shao2022berry}). (2) Alternatively, we can smooth the original loss function by some approximated surrogate functions that are twice differentiable and satisfy Assumption \ref{assumption: RCforIEO}. The surrogate functions can approximate the original function arbitrarily well by Lemma C.1 in \cite{wang2018approximate}.

By Theorem 5.7 in \cite{van2000asymptotic} and Theorem 5.23 in \citet{van2000asymptotic}, which are standard results of $M$-estimation theory, we have the consistency and asymptotic normality result for IEO, and ETO as follows. 

\begin{subtheorem}{proposition}\label{prop: asymptotic for all}

\begin{proposition} [Consistency and asymptotic normality for ETO, Propositions 1.B and 2.B in \cite{elmachtoub2023estimate}] \label{prop: asymptotic for ETO}
Under Assumption \ref{assumption: ETOconsistency}, $\hat{\thete}^{\eto}\xrightarrow{P}\thete^{\kl}$. Under Assumption \ref{assumption: ETOconsistency} and \ref{assumption: RCforETO}, $\sqrt{n} (\hat{\thete}^{\eto}- \thete^{\kl})$ converges in distribution to $\NN_1^\eto$, which is a normal distribution with mean zero and covariance matrix 
\begin{align}
\var_P(\NN_1^\eto):=\rbr{\nabla_{\thete\thete}\mathbb{E}_{P}[\log p_{\thete^{\kl}}(\ran)]}^{-1}
\var_{P}(\nabla_\thete \log p_{\thete^{\kl}}(\ran))(\nabla_{\thete\thete}\mathbb{E}_{P}[\log p_{\thete^{\kl}}(\ran)])^{-1}\label{eqn: cov matrix of ETO}
\end{align}
where $\var_{P}(\nabla_\thete \log p_{\thete^{\kl}}(\ran))$ is the covariance matrix of $\nabla_\thete \log p_{\thete^{\kl}}(\ran)$ under $P$.
Moreover, when $\thete^{\kl}$ corresponds to the ground-truth $P$, i.e., $P_{\thete^{\kl}}=P$,{ and the model has common support near $\thete^{\kl}$ with $\int p_\thete(\ran)d\ran=1$ twice differentiable under the integral,} the covariance matrix \eqref{eqn: cov matrix of ETO} is simplified to the inverse Fisher information $\mathcal{I}_{\thete^{\kl}}^{-1}$, that is,
$$\eqref{eqn: cov matrix of ETO} = \mathcal{I}_{\thete^{\kl}}^{-1}= \rbr{\mathbb{E}_{P}[ (\nabla_\thete \log p_{\thete^{\kl}}(\ran))^\top \nabla_\thete \log p_{\thete^{\kl}}(\ran)]}^{-1}.$$ 
\end{proposition}

\begin{proposition} [Consistency and asymptotic normality for IEO, Propositions 1.C and 2.C in \cite{elmachtoub2023estimate}] \label{prop: asymptotic for IEO}
Under Assumption \ref{assumption: IEOconsistency}, $\hat{\thete}^{\ieo}\xrightarrow{P}\thete^{*}$. Under Assumption \ref{assumption: IEOconsistency} and \ref{assumption: RCforIEO}, $\sqrt{n} (\hat{\thete}^{\ieo}- \thete^*)$ converges in distribution to $\NN_1^\ieo$, which is a normal distribution with mean zero and covariance matrix 
\begin{align*}
\var_P(\NN_1^\ieo):=\nabla_{\thete\thete}\vale_0(\omege_{\thete^*})^{-1}\var_{P}(\nabla_\thete \cost(\omege_{\thete^*},\ran))\nabla_{\thete\thete}\vale_0(\omege_{\thete^*})^{-1}
\end{align*}
where $\var_{P}(\nabla_\thete \cost(\omege_{\thete^*},\ran))$ is the covariance matrix of the cost gradient $\nabla_\thete \cost(\omege_{\thete^*},\ran)$ under $P$.
\end{proposition}
\end{subtheorem}





We list out the following assumption adopted from \citet{elmachtoub2023estimate}.
\begin{assumption}[Smoothness and gradient-expectation interchangeability] \label{SCforh}
Suppose that:
\begin{enumerate}
\item $\vale(\omege, \thete)$ is twice differentiable with respect to $(\omege, \thete)$.

\item The optimal solution $\omege_{\thete}$ to the oracle problem \eqref{eqn: oracle} satisfies that $\omege_{\thete}$ is twice differentiable with respect to $\thete$.

\item Any involved operations of integration (expectation) and differentiation can be interchanged.
Specifically, for any $\thete\in\Theta$,
$$ 
\nabla_\thete \int \nabla_{\omege} \cost(\omege,\ran)^\top   p_\thete(\ran)  d\ran =  \int \nabla_{\omege} \cost(\omege,\ran)^\top \nabla_\thete p_\thete(\ran)  d\ran, $$
$$ 
\int \nabla_{\omege} \cost(\omege,\ran) p_\thete(\ran)  d\ran =  \nabla_\omege \int  \cost(\omege,\ran) p_\thete(\ran) d\ran.$$
\end{enumerate}
\end{assumption}

It is well-known that Berry-Essen-type bounds provide a finite-sample version of the asymptotic normality. Below is an existing theorem of Berry-Esseen bounds for general $M$-estimation in statistical theory. It enhances the classical asymptotic results (e.g., Proposition \ref{prop: asymptotic for all}) by providing a finite-sample performance guarantee.

\begin{lemma}[Berry-Esseen bounds for $M$-estimators, Theorem 3.1 in \cite{shao2022berry})]
    \label{lemma: berry-esseen for M-estimator}
    Suppose the i.i.d. random variables $\ran,\ran_1,...,\ran_n$ follows the distribution $P$. The parameter space of $\zeta$ is {an open convex subset $D$ of $\RR^{d_\zeta}$}.  Suppose that $m(\zeta,\bz)$ is {a finite measurable function of $\bz$}. Let $M(\zeta):=\EE_P(m(\zeta,\ran))$ and 
    \begin{align*}
        \zeta^*:=\argmin_{\zeta}M(\zeta), \qquad \hat{\zeta}_n=\argmin_{\zeta}\frac{1}{n}\sum_{i=1}^nm(\zeta,\ran_i).
    \end{align*}
    {Assume $M$ is finite, differentiation commutes with expectation, both minima are attained in $D$, and $\hat\zeta_n$ is measurable.}
    Moreover, suppose
    \begin{enumerate}[leftmargin=*]
        \item the function $m(\zeta,\cdot)$ is twice {continuously }differentiable with respect to $\zeta$ and there exist constants $\mu>0$, $c_1>0$, $c_2>0$ and two nonnegative functions $K_1, K_2:\cZ\to\RR$ with $\nbr{{K_1(\ran)}}_9\leq c_1$ and $\nbr{{K_2(\ran)}}_4\leq c_2$ such that for any $\zeta$,
        \begin{align*}
            M(\zeta)-M(\zeta^*)&\geq \mu\nbr{\zeta-\zeta^*}^2,\\
            |m(\zeta,\ran)-m(\zeta^*,\ran)|&\leq K_1(\ran)\nbr{\zeta-\zeta^*}, \ \forall \ran\in\cZ,\\
            \nbr{\nabla_{\zeta\zeta}m(\zeta,\ran)-\nabla_{\zeta\zeta}m(\zeta^*,\ran)}&\leq K_2(\ran)\|\zeta-\zeta^*\|,\ \forall \ran\in\cZ.
        \end{align*}
        Moreover, there exists a constant $c_3\geq 0$ and a nonnegative function $K_3:\cZ\to\RR$ such that for any $\ran\in\cZ$,
        \begin{align*}
            {\|\nabla_{\zeta\zeta}m(\zeta^*,\ran)\|\leq K_3(\ran)} \textup{ and } \nbr{K_3(\ran)}_4\leq c_3.
        \end{align*}
        \item Let $\Sigma=\EE[\nabla_\zeta m(\zeta^*,\ran)^\top\nabla_\zeta m(\zeta^*,\ran)]$ and $V=\nabla_{\zeta\zeta}\EE[m(\zeta^*,\ran)]$. 
        Assume that there exist constants $\lambda_1>0$ and $\lambda_2>0$ such that $\lambda_{\min}(\Sigma)\geq \lambda_1$ and $\lambda_{\min}(V)\geq \lambda_2$. 
        Moreover, assume that there exists a constant $c_4>0$ such that $\nbr{\nabla_\zeta m(\zeta^*,\ran)}_4\leq c_4\sqrt{d_\zeta}$.
    \end{enumerate}
    Under the assumption above, let $\cA$ denote all convex {Borel }sets of $\RR^{d_\zeta}$, and we have
    \begin{align*}
        \sup_{A\in\cA}\abr{\PP(\sqrt{n}(\hat{\zeta}_n-\zeta^*)\in A)-\PP(V^{-1}\Sigma^{\frac{1}{2}}\bY_0\in A)}\leq Cd_\zeta^{\frac{9}{4}}n^{-\frac{1}{2}},
    \end{align*}
    where {$\bY_0\sim N(0,\bI_{d_\zeta})$ and }$C>0$ is a constant depending only on $c_1, c_2, c_3, c_4, \mu, \lambda_1, \lambda_2$,
\end{lemma}

Note that since 
$$V^{-1}\Sigma^{\frac{1}{2}}\bY_0 \overset{d}{=} N(0, V^{-1}\Sigma^{\frac{1}{2}} (V^{-1}\Sigma^{\frac{1}{2}})^\top) \overset{d}{=} N(0, V^{-1}\Sigma V^{-1}) \overset{d}{=} (V^{-1}\Sigma V^{-1})^{\frac{1}{2}}\bY_0,$$
the above result will also hold if we replace $\PP(V^{-1}\Sigma^{\frac{1}{2}}\bY_0\in A)$ by $\PP((V^{-1}\Sigma V^{-1})^{\frac{1}{2}}\bY_0\in A)$.

We apply the above Berry-Esseen bounds for $M$-estimators (Lemma \ref{lemma: berry-esseen for M-estimator}) directly to ETO and IEO, which leads to finite-sample performance guarantees in terms of the parameter estimation error.
\begin{subtheorem}{proposition}\label{prop: BEforall}
    \begin{proposition}[Berry-Esseen bounds for ETO]\label{prop: BEforETOparam}
        Under Assumption \ref{assumption: ETOconsistency} and \ref{assumption: RCforETO}, the following holds for any convex {Borel }set $A$:
        \begin{align*}
            |\PP(\sqrt{n}(\hat{\thete}^\eto-\thete^\kl)\in A)-\PP(\bM_1^\eto\bY_0\in A)|\leq C^\eto q^{\frac{9}{4}}n^{-\frac{1}{2}}=:C_{n,q}^\eto,
        \end{align*}
        where $C^\eto>0$ is a constant depending only on $c_1^\eto,c_2^\eto,c_3^\eto,c_4^\eto,\mu^\eto,\lambda_1^\eto,\lambda_2^\eto$.
    \end{proposition}
    \begin{proposition}[Berry-Esseen bounds for IEO]\label{prop: BEforIEOparam}
        Under Assumption \ref{assumption: IEOconsistency} and \ref{assumption: RCforIEO}, the following holds for any convex {Borel }set $A$:
        \begin{align*}
        |\PP(\sqrt{n}(\hat{\thete}^\ieo-\thete^*)\in A)-\PP(\bM_1^\ieo\bY_0\in A)|\leq C^\ieo q^{\frac{9}{4}}n^{-\frac{1}{2}}=:C_{n,q}^\ieo,
        \end{align*}
        where $C^\ieo>0$ is a constant depending only on $c_1^\ieo,c_2^\ieo,c_3^\ieo,c_4^\ieo,\mu^\ieo,\lambda_1^\ieo,\lambda_2^\ieo$.
    \end{proposition}
\end{subtheorem}

\proof[Proof of Proposition \ref{prop: BEforall}]

For ETO, consider $m(\zeta,\ran)=-\log p_{\thete}(\ran)$ with parameter $\zeta=\thete$ and apply Lemma \ref{lemma: berry-esseen for M-estimator}.

For IEO, consider $m(\zeta,\ran)=c(\omege_\thete,\ran)$ with parameter $\zeta=\thete$ and apply Lemma \ref{lemma: berry-esseen for M-estimator}.
\endproof

We give some remarks on Proposition \ref{prop: BEforall}. First, the inequalities in Proposition \ref{prop: BEforall} clearly hold for all sets whose complements are convex.
Second, Proposition \ref{prop: BEforall} can be viewed as a finite-sample extension of Proposition \ref{prop: asymptotic for all} by showing the bound for the finite-sample error to the corresponding normal distributions, which is $O(q^{\frac{9}{4}}n^{-\frac{1}{2}})$. Finally, in general we have $\thete^*\ne \thete^{\kl}$, $\omege_{\thete^*} \ne \omege^*$, $\omege_{\thete^{\kl}} \ne \omege^*$ (unless the model is well-specified). Therefore, the centered parameters in Propositions \ref{prop: BEforETOparam} and \ref{prop: BEforIEOparam} are generally different. This implies that we must tackle the difference of the centered parameters in addition to the distribution error, thus making the comparisons more delicate.

\subsection{Proofs in Section \ref{sec: zeroth-order}}
\proof[Proof of Theorem \ref{thm: higherorder}]
For the zeroth-order term, this result has been established in \citet{elmachtoub2023estimate}. Next, we examine the first-order and second-order terms for ETO and IEO, respectively.

We use Taylor expansion of $R(\omege_\thete)$ at $\thete$ ($\thete = \thete^\kl$ or $\thete^*$), so that
\begin{align*}
&R(\omege_{\hat{\thete}})\\
=&\vale_0(\omege_{\hat{\thete}})-\vale_0(\omege^*)\\
=&\vale_0(\omege_{\hat{\thete}})-\vale_0(\omege_{\thete})+\vale_0(\omege_{\thete})-\vale_0(\omege^*)\\
=&\vale_0(\omege_{\thete})-\vale_0(\omege^*)+\nabla_{\thete}\vale_0(\omege_{\thete})(\hat{\thete}-\thete)+ \frac{1}{2}(\hat{\thete}-\thete)^\top\nabla_{\thete\thete}\vale_0(\omege_{\thete})(\hat{\thete}-\thete)+{o_P(\|\hat{\thete}-\thete\|^2)}.    
\end{align*}
Then we apply the higher-order delta method. 

For IEO, we particularly have $\nabla_{\thete}\vale_0(\omege_{\thete^*})=0$ by the first-order optimality condition. Therefore,  
\begin{align*}
R(\hat{\omege}^{\ieo})
=&\vale_0(\omege_{\thete^{*}})-\vale_0(\omege^*) + \frac{1}{2}(\hat{\thete}^{\ieo}-\thete^{*})^\top\nabla_{\thete\thete}\vale_0(\omege_{\thete^*})(\hat{\thete}^{\ieo}-\thete^{*})+o_P(\|\hat{\thete}^{\ieo}-\thete^{*}\|^2)\\
=&\kappa_0^{\ieo} + \frac{1}{2}(\hat{\thete}^{\ieo}-\thete^{*})^\top\nabla_{\thete\thete}\vale_0(\omege_{\thete^*})(\hat{\thete}^{\ieo}-\thete^{*})+o_P(\|\hat{\thete}^{\ieo}-\thete^{*}\|^2).
\end{align*}
Proposition \ref{prop: asymptotic for IEO} shows that
$$\sqrt{n} (\hat{\thete}^{\ieo}-{\thete^*}) \xrightarrow{d} {\bM_1^{\ieo}} \bm{Y}_0.$$
Thus we have
$$\sqrt{n}(R(\hat{\omege}^{\ieo})-\kappa_0^{\ieo}) \xrightarrow{P} 0.$$
$$n(R(\hat{\omege}^{\ieo})-\kappa_0^{\ieo}) \xrightarrow{d} \mathbb G^{\ieo}.$$

For ETO, in general $\nabla_{\thete}\vale_0(\omege_{\thete^{\kl}})\ne 0$ and thus this term does not vanish. 
\begin{align*}
R(\hat{\omege}^{\eto})
=&\vale_0(\omege_{\thete^{\kl}})-\vale_0(\omege^*) + \nabla_{\thete}\vale_0(\omege_{\thete^{\kl}})(\hat{\thete}^{\eto}-\thete^{\kl})\\
&+ \frac{1}{2}(\hat{\thete}^{\eto}-\thete^{\kl})^\top\nabla_{\thete\thete}\vale_0(\omege_{\thete^{\kl}})(\hat{\thete}^{\eto}-\thete^{\kl})+o_P(\|\hat{\thete}^{\eto}-\thete^{\kl}\|^2).    
\end{align*}

Proposition \ref{prop: asymptotic for ETO} shows that
$$\sqrt{n} (\hat{\thete}^{\eto}-\thete^{\kl}) \xrightarrow{d} {\bM_1^{\eto}} \bm{Y}_0.$$
By the continuous mapping theorem,
$$\sqrt{n} \nabla_{\thete}\vale_0(\omege_{\thete^{\kl}})(\hat{\thete}^{\eto}-\thete^{\kl}) \xrightarrow{d} {\nabla_\thete\vale_0(\omege_{\thete^{\kl}})} {\bM_1^{\eto}} \bm{Y}_0, $$
$$ \frac{1}{2} n (\hat{\thete}^{\eto}-\thete^{\kl})^\top\nabla_{\thete\thete}\vale_0(\omege_{\thete^{\kl}})(\hat{\thete}^{\eto}-\thete^{\kl}) \xrightarrow{d} \mathbb G^{\eto}.$$

Hence we conclude that
$$\sqrt{n} (R(\hat{\omege}^{\eto})-\kappa_0^{\eto}) \xrightarrow{d} {\nabla_\thete\vale_0(\omege_{\thete^{\kl}})} {\bM_1^{\eto}} \bm{Y}_0$$
and
$$n(R(\hat{\omege}^{\eto})-\kappa_0^{\eto}-\nabla_{\thete}\vale_0(\omege_{\thete^{\kl}})(\hat{\thete}^{\eto}-\thete^{\kl})) \xrightarrow{d} \mathbb G^{\eto}.$$
In addition, for any non-decreasing convex function $u:\RR\to\RR$, $$\EE[u({\nabla_\thete\vale_0(\omege_{\thete^{\kl}})} {\bM_1^{\eto}} \bm{Y}_0)]\geq u(\EE({\nabla_\thete\vale_0(\omege_{\thete^{\kl}})} {\bM_1^{\eto}} \bm{Y}_0))={u(0)}.$$
Thus we have ${\nabla_\thete\vale_0(\omege_{\thete^{\kl}})} {\bM_1^{\eto}} \bm{Y}_0\succeq_{\sst} 0$.

\endproof

\proof[Proof of Theorem \ref{thm: Ginmis}]
By the definition of $\GG^\ieo$ and $\GG^\eto$,
\begingroup
orthogonal invariance gives $\GG^a\overset d=\frac12\bm Y_0^\top\bD^a\bm Y_0$ for $a\in\{\eto,\ieo\}$. Set $Z^a=\frac12\bm Y_0^\top\bD^a\bm Y_0$. Then
\begin{align*}
Z^\ieo&=Z^\eto+\frac12\bm Y_0^\top
(\Delta_1+\widetilde{\bD}^\ieo-\widetilde{\bD}^\eto+\Delta_2)\bm Y_0,\\
\Delta_1&:=\widetilde{\bD}^\eto-\bD^\eto,\qquad
\Delta_2:=\bD^\ieo-\widetilde{\bD}^\ieo,\qquad
\Delta:=\Delta_1+\Delta_2.
\end{align*}
\endgroup
Note that in the proof of Lemma 1 in \citet{elmachtoub2023estimate}, it has been shown that
$(\tilde{\bM}_1^{\ieo})^2 \ge (\tilde{\bM}_1^{\eto})^2$ by viewing $P^{\kl}$ as the ground-truth distribution. 
By Lemma \ref{lemma: matrix1}, we have
\begingroup
$$\widetilde{\bD}^{\ieo}\ge\widetilde{\bD}^{\eto}.$$
Hence,
$$\frac12\bm Y_0^\top(\widetilde{\bD}^{\ieo}-\widetilde{\bD}^{\eto})\bm Y_0\ge0.$$
Here $\tilde{\bM}_1^\eto>0$ by the score identity and nonsingular reference Hessian, so also $\tilde{\bM}_1^\ieo>0$; the square-root bounds below therefore apply.
\endgroup
Moreover, by Assumption \ref{assumption: distributionmis}, we claim that
$$\|\Delta_1\| \lesssim {B}_0, \|\Delta_2\| \lesssim {B}_0. $$
To see this, we note the following facts. By the local Lipschitz continuity of matrix inversion, Assumption \ref{assumption: distributionmis} implies
\begin{align*}
   \nbr{(\nabla_{\thete\thete}\mathbb{E}_{P}[\log p_{\thete^{\kl}}(\ran)]))^{-1} - (\nabla_{\thete\thete}\mathbb{E}_{P^{\kl}}[\log p_{\thete^{\kl}}(\ran)]))^{-1}} &\lesssim {B}_0, \\
\nbr{ (\nabla_{\thete\thete}\vale_0(\omege_{\thete^{*}}))^{-1}-(\nabla_{\thete\thete}\vale(\omege_{\thete^\kl},P^{\kl}))^{-1}} &\lesssim {B}_0
\end{align*}
where $\lesssim$ hides the local Lipschitz constant. By using Lemma \ref{lemma: matrix2} and Assumption \ref{assumption: distributionmis}, we have 
\begin{align*}
   &\| \rbr{\nabla_{\thete\thete}\mathbb{E}_{P}[\log p_{\thete^{\kl}}(\ran)]}^{-1}
\var_{P}(\nabla_\thete \log p_{\thete^{\kl}}(\ran))(\nabla_{\thete\thete}\mathbb{E}_{P}[\log p_{\thete^{\kl}}(\ran)])^{-1}\\
&- \rbr{\nabla_{\thete\thete}\mathbb{E}_{P^{\kl}}[\log p_{\thete^{\kl}}(\ran)]}^{-1}
\var_{P^{\kl}}(\nabla_\thete \log p_{\thete^{\kl}}(\ran))(\nabla_{\thete\thete}\mathbb{E}_{P^{\kl}}[\log p_{\thete^{\kl}}(\ran)])^{-1} \| \lesssim {B}_0, \\
   &\| \nabla_{\thete\thete}\vale_0(\omege_{\thete^*})^{-1}\var_{P}(\nabla_\thete \cost(\omege_{\thete^*},\ran))\nabla_{\thete\thete}\vale_0(\omege_{\thete^*})^{-1} \\
&-\nabla_{\thete\thete}\vale(\omege_{\thete^\kl},P^{\kl})^{-1}\var_{P^{\kl}}(\nabla_\thete \cost(\omege_{\thete^\kl},\ran))\nabla_{\thete\thete}\vale(\omege_{\thete^\kl},P^{\kl})^{-1}\| 
\lesssim {B}_0
\end{align*}
In other words,
$$\|(\bM_1^\eto)^2-(\tilde{\bM}_1^\eto)^2\|\lesssim {B}_0, \|(\bM_1^\ieo)^2-(\tilde{\bM}_1^\ieo)^2\|\lesssim {B}_0.$$
By the local Lipschitz continuity of matrix square root, we also have 
$$\|\bM_1^\eto-\tilde{\bM}_1^\eto\|\lesssim {B}_0, \|\bM_1^\ieo-\tilde{\bM}_1^\ieo\|\lesssim {B}_0.$$
By using Lemma \ref{lemma: matrix2} and Assumption \ref{assumption: distributionmis} again, we have
$$\|(\tilde{\bM}_1^{\eto})^\top \nabla_{\thete\thete} 
 \vale_0(\omege_{\thete^{*}}) \tilde{\bM}_1^{\eto}- ({\bM}_1^{\eto})^\top \nabla_{\thete\thete} 
 \vale_0(\omege_{\thete^{\kl}}) {\bM}_1^{\eto}\| \lesssim {B}_0$$
$$\|(\bM_1^{\ieo})^\top\nabla_{\thete\thete}\vale_0(\omege_{\thete^{*}}){\bM_1^{\ieo}} -(\tilde{\bM}_1^{\ieo})^\top \nabla_{\thete\thete} 
 \vale_0(\omege_{\thete^{*}}) \tilde{\bM}_1^{\ieo}\|\lesssim {B}_0$$
 
\begingroup
The eigenvalue perturbation inequality $|\lambda_j(B)-\lambda_j(C)|\le\|B-C\|$ for symmetric $B,C$ gives
\begin{align*}
\|\Delta_1\|&\le\|\widetilde A^\eto-A^\eto\|,\\
\|\Delta_2\|&\le\|A^\ieo-\widetilde A^\ieo\|.
\end{align*}
Thus the preceding matrix bounds apply to the diagonal matrices $\Delta_1,\Delta_2$ defined above.
\endgroup
Hence, we have
$\|\Delta\| = \|\Delta_1+\Delta_2\| \le \|\Delta_1\|+\|\Delta_2\|\lesssim {B}_0$. We thus conclude that there exists a problem-dependent constant $C_\mis$ such that $\nbr{\Delta}\leq C_\mis{B}_0$.
\endproof

\begin{lemma}\label{lemma: matrix2}
Suppose $Q_1$, $Q_2$, $Q_3$, $Q'_1$, $Q'_2$, $Q'_3$ are square matrices. Then we have
$$\|Q_1 Q_2 Q_3 -Q'_1 Q'_2 Q'_3\| \lesssim \left(\max_{i}{\|Q_i-Q_i'\|}\right)  \left(\max_i{(\|Q_i\|, \|Q'_i\|)}\right)^2,$$
$$\|Q_1 Q_2 Q_3 -Q'_1 Q_2 Q'_3\| \lesssim \left(\max_{i}{\|Q_i-Q_i'\|}\right)  \left(\max_i{(\|Q_i\|, \|Q'_i\|)}\right)^2.$$
\end{lemma}

\proof[Proof of Lemma \ref{lemma: matrix2}] 
Note that
\begin{align*}
&\|Q_1 Q_2 Q_3 -Q'_1 Q'_2 Q'_3\|\\ 
\le &  \|(Q_1-Q_1') Q_2 Q_3 + Q'_1 (Q_2-Q_2') Q_3 + Q'_1 Q'_2(Q_3-Q'_3) \|  \\
\le &  \|Q_1-Q_1'\| \|Q_2\| \|Q_3\| + \|Q'_1\| \|Q_2-Q_2'\| \|Q_3\| + \|Q'_1\| \|Q'_2\| \|Q_3-Q'_3\|  \\
\le & 3 \left(\max_{i}{\|Q_i-Q_i'\|}\right)  \left(\max_i{(\|Q_i\|, \|Q'_i\|)}\right)^2
\end{align*}
The other inequality can be derived similarly.
\endproof

\begin{lemma}\label{lemma: matrix1}
Suppose $Q_1$, $Q_2$, and $Q_3$ are all positive definite matrices. If $ Q_1^2\le Q_2^2 $, then we have
\begingroup
$$\lambda_j(Q_1 Q_3 Q_1)\le\lambda_j(Q_2 Q_3 Q_2),\qquad j=1,\ldots,q,$$
where $q$ is the common matrix dimension and $\lambda_j$ denotes the $j$th smallest eigenvalue.
\endgroup
\end{lemma}

\proof[Proof of Lemma \ref{lemma: matrix1}]
We note that
$Q_1 Q_1\le Q_2 Q_2$ implies that
\begingroup
$$Q_3^{\frac12}Q_1^2Q_3^{\frac12}\le Q_3^{\frac12}Q_2^2Q_3^{\frac12}.$$
For $B_i=Q_3^{\frac12}Q_i$, the matrices
$B_i^\top B_i=Q_iQ_3Q_i$ and
$B_iB_i^\top=Q_3^{\frac12}Q_i^2Q_3^{\frac12}$ have the same eigenvalues.
The claimed eigenvalue inequality therefore follows from the variational characterization of eigenvalues.
\endgroup
\endproof

\begingroup
\proof[Proof of Corollary \ref{cor: Ginmis}]
By Theorem \ref{thm: Ginmis},
\[
\bD^{\ieo}-\bD^{\eto}
 =\widetilde{\bD}^{\ieo}-\widetilde{\bD}^{\eto}+\Delta
 \ge(\tau_3-C_\mis B_0)\bI.
\]
Taking the smallest diagonal entry gives $\tau_1\ge\tau_3-C_\mis B_0$.
If $B_0\le\tau_3/C_\mis$, the coupled variables satisfy
\[
Z^{\ieo}-Z^{\eto}
 =\frac12\bm{Y}_0^\top(\bD^{\ieo}-\bD^{\eto})\bm{Y}_0\ge0
\]
almost surely. Their marginal laws are $\mathbb G^{\ieo}$ and $\mathbb G^{\eto}$, proving the claim.
\endproof
\endgroup

\subsection{Proofs in Section \ref{sec: individualbound}}
We first introduce a simple result on the tail upper bound of a multivariate Gaussian distribution.
\begin{lemma}\label{lemma: tailchisquare}
Suppose that $\bX \in \mathbb{R}^q$ is a multivariate Gaussian distribution
$\bX \overset{d}{=}\bM^\top \bY_0$ where $\bM$ is a square matrix and $\bY_0$ is the standard Gaussian distribution. Let $\bv\in \mathbb{R}^q$ be a vector {with $\nbr{\bM\bv}>0$.} For $t\geq 0$, we have
$$\PP(\|\bX\| \ge t)\le 2q  \exp\rbr{-\frac{t^2}{2q\|\bM\|^2}}.$$
$$\PP(\bv^\top \bX \ge t) \le \exp\rbr{-\frac{t^2}{2\|\bM\bv\|^2}}.
$$
For any $s_1\le s_2$, we have
$$\PP(s_1\le \bv^\top \bX \le s_2)
\le \frac{1}{\sqrt{2\pi}} \frac{s_2-s_1}{{\nbr{\bM\bv}}}.$$
\end{lemma}

\proof[Proof of Lemma \ref{lemma: tailchisquare}]
Note that 
\begin{align*}
\PP(\|\bX\| \ge t) =& \PP(\|\bM^\top \bY_0\| \ge t)\\
\le & \PP(\|\bM\| \|\bY_0\| \ge t)\\
\le & \PP \left( \|\bY_0\|^2 \ge \frac{t^2}{\|\bM\|^2} \right)\\
\le & q  \PP \left( (Y_0^{(1)})^2 \ge \frac{t^2}{q\|\bM\|^2} \right)\\
\le & 2q  \exp\rbr{-\frac{t^2}{2q\|\bM\|^2}}.
\end{align*} 
Note that $\bv^\top\bX\overset{d}{=}N(0,\bv^\top\bM^\top\bM\bv)$ so 
$\frac{\bv^\top\bX}{\|\bM\bv\|}\overset{d}{=}N(0,1)$ and thus
$$
\PP(\bv^\top \bX \ge t) = \PP\left( \frac{\bv^\top\bX}{\|\bM\bv\|} \ge \frac{t}{{\|\bM\bv\|}}\right)\leq \exp\rbr{-\frac{t^2}{2\|\bM\bv\|^2}}.
$$
$$
\PP(s_1\le \bv^\top \bX \le s_2) = \PP\left(\frac{s_1}{{\|\bM\bv\|}}\le \frac{\bv^\top\bX}{\|\bM\bv\|} \le \frac{s_2}{{\|\bM\bv\|}}\right)\leq \frac{1}{\sqrt{2\pi}} \frac{s_2-s_1}{{\nbr{\bM\bv}}}.
$$


\endproof

We provide proofs for Proposition \ref{prop: BEforallRegret}. 

\proof[Proof of Proposition \ref{prop: BEforsqrtnETOregret}]
{Define $G_{n,q}^\eto$ as the first supremum below. If $L_2=0$, the remainder is zero and Proposition \ref{prop: BEforall} applies directly. Assume $L_2>0$.}
Since $\nbr{\nabla_{\thete\thete}v_0(\omege_\thete)}\leq L_2$, we have 
    \begin{align*}
        {\left|v_0(\hat{\omege}^\eto)-v_0(\omege_{\thete^\kl})-\nabla_\thete v_0(\omege_{\thete^\kl})(\hat{\thete}^\eto-\thete^\kl)\right|}&\leq \frac{L_2}{2}\nbr{\hat{\thete}^\eto-\thete^\kl}^2.
    \end{align*}
    For ETO, it holds that $\sqrt{n}(v_0(\hat{\omege}^\eto)-v_0(\omege_{{\thete^\kl}}))\xrightarrow{d}\nabla_\thete v_0(\omege_{\thete^\kl})\NN_1^\eto$. Let $\bX_n:=\sqrt{n}(\hat{\thete}^\eto-\thete^\kl)$ and $\bX:=\NN_1^\eto$. Let $f$ be the function with $f(\cdot)=\nabla_\thete v_0(\omege_{\thete^\kl})(\cdot)$. Let  $Y_n={v_0(\hat{\omege}^\eto)-}v_0(\omege_{\thete^\kl})- \nabla_\thete v_0(\omege_{\thete^\kl})(\hat{\thete}^\eto-\thete^\kl)$ with ${|\sqrt{n}Y_n|}\leq L_2\nbr{\bX_n}^2/(2\sqrt{n})$. We have

    \begin{align*}
    &\sup_{{t\in\RR}}\abr{\PP(\sqrt{n}(v_0(\hat{\omege}^\eto)-v_0(\omege_{\thete^\kl}))\geq t)-\PP(\nabla_\thete v_0(\omege_{\thete^\kl})\NN_1^\eto\geq t)}\\
    =&\sup_{{t\in\RR}}\abr{ \PP(f(\bX_n)+\sqrt{n}Y_n\geq t)- \PP(f(\bX)\geq t)}\\
    \leq&\sup_{{t\in\RR}}\abr{ \PP(f(\bX_n)+\sqrt{n}Y_n\geq t)- \PP(f(\bX_n)\geq t)}+\sup_{{t\in\RR}}\abr{ \PP(f(\bX_n)\geq t)- \PP(f(\bX)\geq t)}.
\end{align*}
On one hand,
    \begin{align*}
        \sup_{{t\in\RR}}| \PP(f(\bX_n)\geq t)- \PP(f(\bX)\geq t)|&=\sup_{{t\in\RR}}\abr{ \PP(\bX_n\in \underbrace{f^{-1}([t,\infty))}_{\textup{convex}})- \PP(\bX\in \underbrace{f^{-1}([t,\infty))}_{\textup{convex}})}\le C_{n,q}^\eto.
    \end{align*}
On the other hand,
\begin{align*}
   &| \PP(f(\bX_n)+\sqrt{n}Y_n\geq t)-\PP(f(\bX_n)\geq t)|\\
   \leq & \PP(f(\bX_n)+{\sqrt{n}}Y_n\geq t, f(\bX_n)<t)+\PP(f(\bX_n)+\sqrt{n}Y_n< t, f(\bX_n)\geq t)\\
   \leq & 2\PP\rbr{|\sqrt{n}Y_n|{\geq}|t-f(\bX_n)|}\\
   \leq & 2\PP\rbr{\frac{L_2}{2\sqrt{n}}\|\bX_n\|^2{\geq}|t-f(\bX_n)|}.
\end{align*}
Note that for any $\varepsilon>0$,
\begin{align*}
    &\PP\rbr{\frac{L_2}{2\sqrt{n}}\|\bX_n\|^2{\geq}|t-f(\bX_n)|}\\
    =&\PP\rbr{\frac{L_2}{2\sqrt{n}}\|\bX_n\|^2{\geq}|t-f(\bX_n)|,|t-f(\bX_n)|>\varepsilon}+\PP\rbr{\frac{L_2}{2\sqrt{n}}\|\bX_n\|^2{\geq}|t-f(\bX_n)|,|t-f(\bX_n)|\leq\varepsilon}\\
    \leq&\PP\rbr{\frac{L_2}{2\sqrt{n}}\|\bX_n\|^2>\varepsilon}+\PP\rbr{|t-f(\bX_n)|\leq\varepsilon}.
\end{align*}
Note that
\begin{align*}
    \PP(|t-f(\bX_n)|\leq\varepsilon)&=\PP(\bX_n\in f^{-1}({(-\infty,t+\varepsilon]}))-\PP(\bX_n\in f^{-1}(-\infty,t-\varepsilon))\\
    &\le \PP(\bX\in f^{-1}({(-\infty,t+\varepsilon]}))-\PP(\bX\in f^{-1}(-\infty,t-\varepsilon))+2C_{n,q}^\eto\\
    &=\PP(t-\varepsilon\leq f(\bX)\leq t+\varepsilon)+2C_{n,q}^\eto\\
    &\le \frac{2\varepsilon}{\sqrt{2\pi}\nbr{\nabla_\thete v_0(\omege_{\thete^\kl})\bM_1^\eto}}+2C_{n,q}^\eto.
\end{align*}
where the last inequality follows from Lemma \ref{lemma: tailchisquare}.


Moreover, by Lemma \ref{lemma: tailchisquare},
\begin{align*}
    P\rbr{\frac{L_2}{2\sqrt{n}}\|\bX_n\|^2>\varepsilon}&\le P\rbr{\frac{L_2}{2\sqrt{n}}\|\bX\|^2>\varepsilon}+C_{n,q}^\eto\\
    &\leq 2q\exp\rbr{-\frac{\sqrt{n}\varepsilon}{q\nbr{\bM_1^\eto}^2L_2}}+C_{n,q}^\eto.
\end{align*}



Taking $\varepsilon=\nbr{\bM_1^\eto}^2L_2 q n^{-\frac{1}{2}}{\log(en)}$, we have
\begin{align*}
P\rbr{\frac{L_2}{2\sqrt{n}}\|\bX_n\|^2>\varepsilon}\le 2q n^{-1}+C_{n,q}^\eto
\end{align*}
and 
\begin{align*}
\PP\left(\frac{L_2}{2\sqrt{n}}\|\bX_n\|^2{\geq}|t-f(\bX_n)|\right)\lesssim \frac{\nbr{\bM_1^\eto}^2L_2}{\nbr{\nabla_\thete v_0(\omege_{\thete^\kl})\bM_1^\eto}} q n^{-\frac{1}{2}}{\log(en)} + q n^{-1}+C_{n,q}^\eto
\end{align*}
Combining things together, we have
\begin{align*}
&\sup_{{t\in\RR}}\abr{\PP(\sqrt{n}(v_0(\hat{\omege}^\eto)-v_0(\omege_{\thete^\kl}))\geq t)-\PP(\nabla_\thete v_0(\omege_{\thete^\kl})\NN_1^\eto\geq t)}\\
\lesssim & \frac{\nbr{\bM_1^\eto}^2L_2}{\nbr{\nabla_\thete v_0(\omege_{\thete^\kl})\bM_1^\eto}} q n^{-\frac{1}{2}}{\log(en)} + q n^{-1}+C_{n,q}^\eto.
\end{align*}

\endproof

\proof[Proof of Proposition \ref{prop: BEforsqrtnIEOregret}]
{Here $L_2>0$ by the positive definite expected Hessian in Assumption \ref{assumption: RCforIEO}. Let $G_{n,q,t}^\ieo$ be the probability on the left below. The convex-set bound applies to the complement of the open ball.}
Since $\nbr{\nabla_{\thete\thete}v_0(\omege_\thete)}\leq L_2$, we have
\begin{align*}
v_0(\hat{\omege}^\ieo)-v_0(\omege_{\thete^*})&\leq \frac{L_2}{2}\nbr{\hat{\thete}^\ieo-\thete^*}^2.
\end{align*}
For IEO, it holds that $\sqrt{n}(v_0(\hat{\omege}^\ieo)-v_0(\omege_{\thete^*}))\xrightarrow{P}0$. Let $\bX_n:=\sqrt{n}(\hat{\thete}^\ieo-\thete^*)$ and $\bX:=\NN_1^\ieo$. Following a similar argument as in Proposition \ref{prop: BEforsqrtnETOregret}, we obtain that
\begin{align*}
\PP(\sqrt{n}(v_0(\hat{\omege}^\ieo)-v_0(\omege_{\thete^*}))\geq t)
&\leq\PP(\sqrt{n}\frac{L_2}{2}\nbr{\hat{\thete}^\ieo-\thete^*}^2\geq t)\\
&=\PP(\nbr{\bX_n}^2\geq \frac{2\sqrt{n}t}{L_2})\\
&\leq \PP(\nbr{\bX}^2\geq \frac{2\sqrt{n}t}{L_2})+C_{n,q}^\ieo\\
&\leq 2q\exp\rbr{-\frac{\sqrt{n}t}{qL_2\nbr{\bM_1^\ieo}^2}}+C_{n,q}^\ieo.
\end{align*}
\endproof

Next, without loss of generality, we mainly focus on Proposition \ref{prop: BEfornIEOregret}, since the techniques are almost the same for ETO in Proposition \ref{prop: BEfornETOregret}. We first introduce the following two technical lemmas in optimization and probability literature.

\begin{lemma}
    [Lemma 1.2.4 in \cite{nesterov2013introductory}]\label{lemma: third order} For a twice differentiable function ${f:U\to\R}${, where $U\subset\RR^n$ is open and convex}, if it has $M$-Lipschitz Hessian, i.e.,
    \begin{align*}
        \|\nabla^2f(\bx)-\nabla^2f(\by)\|\leq M\|\bx-\by\|,
    \end{align*}
    then
    \begin{align*}
        \abr{f(\by)-f(\bx)-\nabla f(\bx)(\by-\bx)-\frac{1}{2}(\by-\bx)^\top\nabla^2f(\bx)(\by-\bx)}\leq\frac{M}{6}\nbr{\by-\bx}^3.
    \end{align*}
\end{lemma}
\begin{lemma}[Quadratic forms of random variables \citep{muirhead1982quadratic}]
    \label{lemma: quadratic form of rv} 
    Let $\bX$ be $q\times 1$ random vector with mean value $\bmu$ and {positive definite covariance matrix $\bSigma$}. Then for a $q\times q$ symmetric nonsingular matrix $\bA$, the quadratic form $Q(\bX):=\bX^\top\bA\bX$ has the following representation:
    \begin{align*}
        Q(\bX)=\bX^\top \bA\bX=\sum_{j=1}^q\lambda_j(U_j+b_j)^2,
    \end{align*}
    for some $\lambda_1,...,\lambda_q\in\RR$, some fixed vector $\bb\in\RR^q$ and some random vector $\bU$ with $\EE(\bU)=0$ and $\cov(\bU)=\bI$.
\end{lemma}
\proof[Proof of Lemma \ref{lemma: quadratic form of rv}]

By denoting $\bY=\bSigma^{-\frac{1}{2}}\bX$ and $\bZ=\bY-\bSigma^{-\frac{1}{2}}\bmu$, we have
\begin{align*}
    \EE(\bY)=\bSigma^{-\frac{1}{2}}\bmu,\qquad &{\cov(\bY)}=\bSigma^{-\frac{1}{2}}\cov(\bX)\bSigma^{-\frac{1}{2}}=\bSigma^{-\frac{1}{2}}\bSigma\bSigma^{-\frac{1}{2}}=\bI,\\
    \EE(\bZ)=\bm{0},\qquad&\cov(\bZ)=\bI,
\end{align*}
and
\begin{align*}
    Q(\bX)=\bX^\top\bA\bX=\bY^\top\bSigma^{\frac{1}{2}}\bA\bSigma^{\frac{1}{2}}\bY={(\bZ+\bSigma^{-\frac{1}{2}}\bmu)^\top}\bSigma^{\frac{1}{2}}\bA\bSigma^{\frac{1}{2}}(\bZ+\bSigma^{-\frac{1}{2}}\bmu).
\end{align*}
Let $\bP$ be a $q\times q$ orthonormal matrix that diagonalizes $\bSigma^{\frac{1}{2}}\bA\bSigma^{\frac{1}{2}}$ with 
\begin{align*}
    \bP^\top\bSigma^{\frac{1}{2}}\bA\bSigma^{\frac{1}{2}}\bP=\diag(\lambda_1,...,\lambda_q),\qquad \bP^\top\bP=\bP\bP^\top=\bI,
\end{align*}
where $\lambda_1,...,\lambda_q$ are the eigenvalues of $\bSigma^{\frac{1}{2}}\bA\bSigma^{\frac{1}{2}}$. Letting $\bU=\bP^\top\bZ$, we have
\begin{align*}
    {\bZ=\bP\bU}, \qquad, \EE(\bU)=\bm{0},\qquad\cov(\bU)=\bI.
\end{align*}
Then by setting $\bb=\bP^\top{\bSigma^{-\frac{1}{2}}}\bmu$,
\begin{align*}
    Q(\bX)&={(\bZ+\bSigma^{-\frac{1}{2}}\bmu)^\top}\bSigma^{\frac{1}{2}}\bA\bSigma^{\frac{1}{2}}(\bZ+\bSigma^{-\frac{1}{2}}\bmu)=(\bU+\bb)^\top\bP^\top\bSigma^{\frac{1}{2}}\bA\bSigma^{\frac{1}{2}}\bP (\bU+\bb)\\
    &=(\bU+\bb)^\top\diag(\lambda_1,...,\lambda_q) (\bU+\bb)=\sum_{j=1}^q\lambda_j(U_j+b_j)^2.
\end{align*}
\endproof

In particular, when $\bX$ is Gaussian, then $Q(\bX)$ is a linear combination of independent chi-square variables when $\bmu=\bm{0}$ and of noncentral chi-square variables when $\bmu\not=\bm{0}$. More generally, for the general quadratic form with nonsingular symmetric matrix $\bA${ and scalar $c$},
\begin{align*}
    Q(\bX)&=\frac{1}{2}\bX^\top\bA\bX+\bd^\top\bX+{c}=\frac{1}{2}({\bX+\bA^{-1}\bd})^\top\bA({\bX+\bA^{-1}\bd})+{c}-\frac{1}{2}\bd^\top\bA^{-1}\bd\\
    &=\sum_{j=1}^q\lambda_j(U_j+b_j)^2+{c}-\frac{1}{2}\bd^\top\bA^{-1}\bd,
\end{align*}
for some $\lambda_1,...,\lambda_q\in\RR$, some fixed vector $\bb\in\RR^q$ and some random vector $\bU$ with $\EE(\bU)=0$ and $\cov(\bU)=\bI$.

\proof[Proof of Proposition \ref{prop: BEfornIEOregret}]

For simplicity, let $\bX_n:=\sqrt{n}(\hat{\thete}^\ieo-\thete^*)$, $\bX:=\NN_1^\ieo$ and denote the quadratic function $f:\RR^q\to\RR$, $f(\bx)=\frac{1}{2}\bx^\top\nabla_{\thete\thete}v_0(\omege_{\thete^*})\bx$. In this case, $\GG^\ieo=f(\bX)$. Before going to the proof, we point our three facts.

\textbf{Claim 1.} For all $\thete\in\Theta$, $v_0$ satisfies:
    \begin{align}\label{eqn: third order IEO}
        \abr{v_0(\omege_\thete)-v_0(\omege_{\thete^*})-\frac{1}{2}({\thete}-\thete^*)^\top\nabla_{\thete\thete}v_0(\omege_{\thete^*})({\thete}-\thete^*)}\leq \frac{L_1}{6}\|{\thete}-\thete^*\|^3.
    \end{align}
{This follows from Assumption \ref{assumption: Lipschitz Hessian of v0}, $\nabla_\thete v_0(\omege_{\thete^*})=0$, and Lemma \ref{lemma: third order}.}
    
\textbf{Claim 2.} We have $\EE\|\bX\|^r_2<\infty$ for all $r\geq 2$. To see this, For each fixed vector $\bx\in\RR^q$, and any $2\leq r<\infty$, by H\"oder's inequality, we have $\|\bx\|_2\leq q^{\frac{1}{2}-1/r}\|\bx\|_r$. Hence,
\begin{align*}
    \EE[\|\bX\|_2^r]&\leq \EE[(q^{\frac{1}{2}-1/r}\|\bX\|_r)^r]=q^{r/2-1}\EE\|\bX\|^r_r=q^{r/2-1}\EE[\sum_{i=1}^q|\bX_i|^r]\leq q^{r/2-1}\sum_{i=1}^q\EE|\bX_i|^r\\
    &\leq q^{r/2-1}\sum_{i=1}^q(K_i\sqrt{r})^r\leq q^{r/2}(\max_i K_i\sqrt{r})^{r}{=(\max_{i\in[q]} K_i\sqrt{qr})^{r}},
\end{align*}
where the second last inequality holds because each component of $\bX$, say $\bX^{(i)}$, is a Gaussian distribution with variance proxy $K_i^2$, satisfying the moment property (Proposition 2.5.2 in \cite{vershynin2018high}).

\textbf{Claim 3.} From Lemma \ref{lemma: quadratic form of rv}, $\GG^\ieo$ is the linear combination of independent Chi-squared distributions with degree of freedom $1$. Moreover, $\GG^\ieo$ has bounded density when $q\geq 2$. More precisely, 
\begin{align*}
    \GG^\ieo=\sum_{i=1}^q\lambda_i\chi^2(1),
\end{align*}
where ${\lambda_1\ge\cdots\ge\lambda_q>0}$ are eigenvalues of the matrix $\frac{1}{2}\bM_1^\ieo \nabla_{\thete\thete}v_0(\omege_{\thete^*}) \bM_1^\ieo$. When $q=1$, $\GG^\ieo$ is a scaled version of the chi-square distribution with a degree of freedom $1$. When $q\geq 2$, by Theorem 1 of \cite{bobkov2020two}, the density of $\GG^\ieo$, say $p_{\GG^\ieo}(x)$, satisfies
\begin{align*}
    \frac{1}{4e^2\sqrt{2\pi}}\left(\rbr{\sum_{i=1}^q\lambda_i^2}\rbr{\sum_{i=2}^q\lambda_i^2}\right)^{-\frac{1}{4}}\leq\sup_{x>0}p_{\GG^\ieo}(x)\leq\frac{2}{\sqrt{\pi}}\left(\rbr{\sum_{i=1}^q\lambda_i^2}\rbr{\sum_{i=2}^q\lambda_i^2}\right)^{-\frac{1}{4}}.
\end{align*}

{Define $D_{n,q}^\ieo$ as the supremum below. If $L_1=0$, the remainder vanishes and the convex-set bound gives the result directly. Assume $L_1>0$.}
From the assumption, we define $Y_n$ by
\begin{align*}
    Y_n:=v_0(\omege_{{\hat{\thete}^\ieo}})-v_0(\omege_{\thete^*})-\frac{1}{2}(\hat{\thete}^\ieo-\thete^*)^\top\nabla_{\thete\thete}v_0(\omege_{\thete^*})(\hat{\thete}^\ieo-\thete^*)\leq\frac{L_1}{6}\|\hat{\thete}^\ieo-\thete^*\|^3,
\end{align*}
with $|nY_n|\leq \frac{L_1}{6\sqrt{n}}\|\bX_n\|^3$. For all $t\geq 0$, 
\begin{align*}
    &\sup_{t\geq 0}\abr{ \PP(nv_0(\hat{\omege}^\ieo)-nv_0(\omege_{\thete^*})\ge t)- \PP(\GG^{\ieo}\ge t)}\\
    =&\sup_{t\geq 0}\abr{P\rbr{\frac{n}{2}(\hat{\thete}^\ieo-\thete^*)^\top\nabla_{\thete\thete}v_0(\omege_{\thete^*})(\hat{\thete}^\ieo-\thete^*)+nY_n{\ge} t}- \PP(\GG^{\ieo}\ge t)}\\
    =&\sup_{t\geq 0}\abr{ \PP(f(\bX_n)+nY_n\ge t)- \PP(f(\bX)\ge t)}\\
    {\leq}&\sup_{t\geq 0}\abr{ \PP(f(\bX_n)+nY_n\ge t)- \PP(f(\bX_n)\ge t)}+\sup_{t\geq 0}\abr{ \PP(f(\bX_n)\ge t)- \PP(f(\bX)\ge t)}
\end{align*}
    From Proposition \ref{prop: BEforIEOparam},
    \begin{align*}
        \sup_{t\geq 0}| \PP(f(\bX_n)\ge t)- \PP(f(\bX)\ge t)|&=\sup_{t\geq 0}\abr{ \PP(\bX_n\in \underbrace{f^{-1}({(-\infty,t)})}_{\textup{convex}})- \PP(\bX\in \underbrace{f^{-1}({(-\infty,t)})}_{\textup{convex}})}\le C_{n,q}^\ieo.
    \end{align*}
On the other hand,
\begin{align*}
   &| \PP(f(\bX_n)+nY_n\ge t)-\PP(f(\bX_n)\ge t)|\\
   \leq & \PP(f(\bX_n)+nY_n\ge t, f(\bX_n)<t)+\PP(f(\bX_n)+nY_n<t, f(\bX_n)\ge t)\\
   \leq & 2\PP\rbr{|nY_n|{\geq}|t-f(\bX_n)|}\\
   \leq & 2\PP\rbr{\frac{L_1}{6\sqrt{n}}\|\bX_n\|^3{\geq}|t-f(\bX_n)|}.
\end{align*}
Note that for any $\varepsilon>0$,
\begin{align*}
    &\PP\rbr{\frac{L_1}{6\sqrt{n}}\|\bX_n\|^3{\geq}|t-f(\bX_n)|}\\
    =&\PP\rbr{\frac{L_1}{6\sqrt{n}}\|\bX_n\|^3{\geq}|t-f(\bX_n)|,|t-f(\bX_n)|>\varepsilon}+\PP\rbr{\frac{L_1}{6\sqrt{n}}\|\bX_n\|^3{\geq}|t-f(\bX_n)|,|t-f(\bX_n)|\leq\varepsilon}\\
    \leq&\PP\rbr{\frac{L_1}{6\sqrt{n}}\|\bX_n\|^3>\varepsilon}+\PP\rbr{|t-f(\bX_n)|\leq\varepsilon}.
\end{align*}

When $q\geq 2$,  from Proposition \ref{prop: BEforIEOparam} and the fact that $\GG^\ieo$ has a bounded density

\begin{align*}
    \PP(|t-f(\bX_n)|\leq\varepsilon)&=\PP(\bX_n\in f^{-1}({(-\infty,t+\varepsilon]}))-\PP(\bX_n\in f^{-1}(-\infty,t-\varepsilon))\\
    &{\le} \PP(\bX\in f^{-1}({(-\infty,t+\varepsilon]}))-\PP(\bX\in f^{-1}(-\infty,t-\varepsilon))+{2}C_{n,q}^\ieo\\
    &=\PP(t-\varepsilon\leq\GG^\ieo\leq t+\varepsilon)+{2}C_{n,q}^\ieo\\
&\lesssim\left(\rbr{\sum_{i=1}^q\lambda_i^2}\rbr{\sum_{i=2}^q\lambda_i^2}\right)^{-\frac{1}{4}}\varepsilon+C_{n,q}^\ieo.
\end{align*}





Moreover, by Lemma \ref{lemma: tailchisquare},
\begin{align*}
    \PP\rbr{\frac{L_1}{6\sqrt{n}}\|\bX_n\|^3>\varepsilon}&\le \PP\rbr{\frac{L_1}{6\sqrt{n}}\|\bX\|^3>\varepsilon}+C_{n,q}^\ieo\\
    &\leq 2q\exp\rbr{-\frac{1}{2q\nbr{\bM_1^\ieo}^2} \left(\frac{6\sqrt{n} \varepsilon}{L_1}\right)^{2/3}}+C_{n,q}^\ieo.
\end{align*}

Taking $\varepsilon=\frac{L_1}{6\sqrt{n}} (2q\nbr{\bM_1^\ieo}^2 {\log(en)})^{3/2}$, we have
\begin{align*}
    P\rbr{\frac{L_1}{6\sqrt{n}}\|\bX_n\|^3>\varepsilon}&\le 2q n^{-1}+C_{n,q}^\ieo.
\end{align*}
Hence,
\begin{align*}
    \PP\left(\frac{L_1}{6\sqrt{n}}\|\bX_n\|^3{\geq}|t-f(\bX_n)|\right)\lesssim \rbr{\rbr{\sum_{i=1}^q\lambda_i^2}\rbr{\sum_{i=2}^q\lambda_i^2}}^{-\frac{1}{4}}L_1 \nbr{\bM_1^\ieo}^3 q^{\frac{3}{2}} ({\log(en)})^{\frac{3}{2}} n^{-\frac{1}{2}}+q n^{-1}+C_{n,q}^\ieo.
\end{align*}
Combining things together, we have 
\begin{align*}
    &\sup_{t\geq 0}|\PP(nv_0(\hat{\omege}^\ieo)-nv_0(\omege_{\thete^*})\ge t)-\PP(\GG^{\ieo}\ge t)|\\
    \lesssim &\rbr{\rbr{\sum_{i=1}^q\lambda_i^2}\rbr{\sum_{i=2}^q\lambda_i^2}}^{-\frac{1}{4}}L_1\nbr{\bM_1^\ieo}^3q^{3/2}({\log(en)})^{\frac{3}{2}} n^{-\frac{1}{2}}+q n^{-1}+C_{n,q}^\ieo,
\end{align*}
where $``\lesssim''$ hides the constants that is independent of $n$ and $q$.


For $q=1$, we have $\GG^\ieo=\lambda_1\chi^2(1)$. {The same convex-set comparison bounds the empirical interval probability by the Gaussian interval probability plus $2C_{n,1}^\ieo$.} For any $t,\varepsilon>0$,
\begin{align*}
    \PP(t-\varepsilon\leq\GG^\ieo\leq t+\varepsilon)=\PP\rbr{\frac{1}{\lambda_1}(t-\varepsilon)\leq\chi^2(1)\leq \frac{1}{\lambda_1}(t+\varepsilon)}=2\PP\rbr{N(0,1)\in\sbr{\sqrt{\frac{(t-\varepsilon)_+}{\lambda_1}},\sqrt{\frac{t+\varepsilon}{\lambda_1}}}}.
\end{align*}
If $t<\varepsilon$, 
\begin{align*}
   \PP\rbr{N(0,1)\in\sbr{\sqrt{\frac{(t-\varepsilon)_+}{\lambda_1}},\sqrt{\frac{t+\varepsilon}{\lambda_1}}}}\leq \PP\rbr{0\leq N(0,1)\leq \sqrt{\frac{2\varepsilon}{\lambda_1}}}\lesssim \sqrt{\varepsilon/\lambda_1}.
\end{align*}
If $t\geq \varepsilon$, $\sqrt{(t+\varepsilon)/\lambda_1}-\sqrt{(t-\varepsilon)/\lambda_1}\leq \sqrt{2\varepsilon/\lambda_1}$. Hence,
\begin{align*}
    \PP\rbr{N(0,1)\in\sbr{\sqrt{\frac{t-\varepsilon}{\lambda_1}},\sqrt{\frac{t+\varepsilon}{\lambda_1}}}}\leq \PP(0\leq N(0,1)\leq\sqrt{\frac{2\varepsilon}{\lambda_1}})\lesssim \sqrt{\varepsilon/\lambda_1}.
\end{align*}
Taking $\varepsilon=\frac{L_1}{6\sqrt{n}} (2\nbr{\bM_1^\ieo}^2 {\log(en)})^{3/2}$ (where $q=1$), we have 
$$\PP\left(\frac{L_1}{6\sqrt{n}}\|\bX_n\|^3{\geq}|t-f(\bX_n)|\right)\lesssim \frac{1}{\sqrt{\lambda_1}} L_1^{\frac{1}{2}} \nbr{\bM_1^\ieo}^{\frac{3}{2}} ({\log(en)})^{\frac{3}{4}} n^{-\frac{1}{4}}+ n^{-1} + C_{n,1}^\ieo$$

Combining things together, we have 
\begin{align*}
\sup_{t\geq 0}|\PP(nv_0(\hat{\omege}^\ieo)-nv_0(\omege_{\thete^*})\geq t)-\PP(\GG^{\ieo}\geq t)|\lesssim \frac{1}{\sqrt{\lambda_1}} L_1^{\frac{1}{2}} \nbr{\bM_1^\ieo}^{\frac{3}{2}} ({\log(en)})^{\frac{3}{4}} n^{-\frac{1}{4}}+ n^{-1} + C_{n,1}^\ieo.
\end{align*}

\endproof
We can immediately prove Corollary \ref{cor: Fast rate, Berry Esseen} by Proposition \ref{prop: BEfornIEOregret}.
\proof[Proof of Corollary \ref{cor: Fast rate, Berry Esseen}]

We show the result for $q\geq 2$. The case of $q=1$ can be proven similarly. By Proposition \ref{prop: BEfornIEOregret}, {taking thresholds decreasing to $t$ and using continuity of $\GG^\ieo$,}  there exists a problem dependent $C_{\textup{prob}}$ such that
    \begin{align*}
        \PP(nv_0(\hat{\omege}^\ieo)-nv_0(\omege_{\thete^*})\leq t)\geq \PP(\GG^{\ieo}\leq t)-C_{\textup{prob}}({\log(en)})^{\frac{3}{2}} n^{-\frac{1}{2}}.
        \end{align*}
When $n$ is larger than a threshold such that $C_{\textup{prob}}({\log(en)})^{\frac{3}{2}} n^{-\frac{1}{2}}\leq \varepsilon/2$, since $\PP(\GG^{\ieo}\leq F_{\GG^{\ieo}}^{-1}(1-\varepsilon/2))=1-\varepsilon/2$, we have 
\begin{align*}
\PP(\GG^{\ieo}\leq F_{\GG^{\ieo}}^{-1}(1-\varepsilon/2))-C_{\textup{prob}}({\log(en)})^{\frac{3}{2}} n^{-\frac{1}{2}}=1-\varepsilon/2-C_{\textup{prob}}({\log(en)})^{\frac{3}{2}} n^{-\frac{1}{2}}\geq 1-\varepsilon.
\end{align*}
In conclusion, when $n$ satisfies $C_{\textup{prob}}({\log(en)})^{\frac{3}{2}} n^{-\frac{1}{2}}\leq \varepsilon/2$, with probability at least $1-\varepsilon$, $R(\hat{\omege}^\ieo)\leq \kappa_0^\ieo+\frac{F_{\GG^{\ieo}}^{-1}(1-\varepsilon/2)}{n}$.
\endproof

We can use the same techniques of Proposition \ref{prop: BEfornIEOregret} to build the proof of Proposition \ref{prop: BEfornETOregret}.
\proof[Proof of Proposition \ref{prop: BEfornETOregret}]
 The idea is similar to Proposition \ref{prop: BEfornIEOregret}. First,  we note by Assumption \ref{assumption: Lipschitz Hessian of v0} and Lemma \ref{lemma: third order} that  for all $\thete\in\Theta$, $v_0$ satisfies:
    \begin{align}\label{eqn: third order ETO}
        |v_0(\omege_\thete)-v_0(\omege_{\thete^\kl})-\nabla_\thete v_0(\omege_{\thete^\kl})({\thete}-\thete^\kl)-\frac{1}{2}({\thete}-\thete^\kl)^\top\nabla_{\thete\thete}v_0(\omege_{\thete^\kl})({\thete}-\thete^\kl)|\leq \frac{L_1}{6}\|{\thete}-\thete^\kl\|^3.
    \end{align}
    For simplicity, let $\bX_n:=\sqrt{n}(\hat{\thete}^\eto-\thete^\kl)$, $\bX:=\NN_1^\eto$ and denote the quadratic function $f:\RR^q\to\RR$, $f(\bx)=\frac{1}{2}\bx^\top\nabla_{\thete\thete}v_0(\omege_{\thete^\kl})\bx$. By the assumption, we know that ${f}$ is a convex function and has a convex sub-level set. The remaining analysis is almost the same by setting 
\begin{align*}
    Y_n:=&v_0(\hat{\omege}^{\eto})-v_0(\omege_{\thete^\kl})-\nabla_\thete v_0(\omege_{\thete^\kl})({\hat{\thete}^\eto}-\thete^\kl)-\frac{1}{2}(\hat{\thete}^\eto-\thete^\kl)^\top\nabla_{\thete\thete}v_0(\omege_{\thete^\kl})(\hat{\thete}^\eto-\thete^\kl)\\
    \leq&\frac{L_1}{6}\|\hat{\thete}^\eto-\thete^\kl\|^3,
\end{align*}
with $|nY_n|\leq \frac{L_1}{6\sqrt{n}}\|\bX_n\|^3$. The remaining analysis is exactly the same as Proposition \ref{prop: BEfornIEOregret}. {Define $D_{n,q}^\eto$ as the corresponding supremum of the absolute tail-probability difference.}

\endproof


\subsection{Proofs in Section \ref{sec: finite-sample}}

\proof[Proof of Theorem \ref{thm: finitecomparemis2}]

Case 1: $t\le \kappa_0^{\ieo}$. In this case, $R(\hat{\omege}^{\ieo})\ge R(\omege_{\thete^*}) = \kappa_0^{\ieo}\ge t$ and $R(\hat{\omege}^{\eto})\ge R(\omege_{\thete^*}) = \kappa_0^{\ieo}\ge t$ as any realization of $\hat{\omege}^{\ieo}$ or $\hat{\omege}^{\eto}$ should have larger regret than $\omege_{\thete^*}$. So
$$\PP(R(\hat{\omege}^{\eto})\ge t) - \PP( R(\hat{\omege}^{\ieo})\ge t)=1-1=0.$$

Case 2. For any $t > \kappa_0^\ieo$, by Proposition \ref{prop: BEforsqrtnIEOregret},
\begin{align*}
    \PP(R(\hat{\omege}^\ieo)\geq t)&=\PP\rbr{v_0(\hat{\omege}^\ieo)-v_0(\omege_{\thete^*})\geq t-\kappa_0^\ieo}\\
    &=\PP\rbr{\sqrt{n}\rbr{v_0(\hat{\omege}^\ieo)-v_0(\omege_{\thete^*})}\geq \sqrt{n}(t-\kappa_0^\ieo)}\\
    &\le G^\ieo_{n,q,\sqrt{n}(t-\kappa_0^\ieo)}.
\end{align*}
On the other hand, by Proposition \ref{prop: BEforsqrtnETOregret},
\begin{align*}
&\PP(R(\hat{\omege}^\eto)\geq t)-1\\
= &-\PP(R(\hat{\omege}^\eto){<} t)\\
= & - \PP(v_0(\hat{\omege}^\eto)-v_0(\omege_{\thete^\kl}){<}t-\kappa_0^\eto)\\
= & - \PP(\sqrt{n}(v_0(\hat{\omege}^\eto)-v_0(\omege_{\thete^\kl})){<}\sqrt{n}(t-\kappa_0^\eto))\\
\ge & - \PP(\nabla_\thete v_0(\omege_{\thete^\kl})\NN_1^\eto\le \sqrt{n}(t-\kappa_0^\eto))-G_{n,q}^\eto.
\end{align*}
Thus,
\begin{align*}
&\PP(R(\hat{\omege}^\ieo)\geq t) - \PP(R(\hat{\omege}^\eto)\geq t) \\
\le & -1 + \PP(\nabla_\thete v_0(\omege_{\thete^\kl})\NN_1^\eto\le \sqrt{n}(t-\kappa_0^\eto))+ G^\ieo_{n,q,\sqrt{n}(t-\kappa_0^\ieo)}+ G_{n,q}^\eto
\end{align*}
or
\begin{align*}
& \PP(R(\hat{\omege}^\eto)\geq t)-\PP(R(\hat{\omege}^\ieo)\geq t) \\
\ge & 1 - \PP(\nabla_\thete v_0(\omege_{\thete^\kl})\NN_1^\eto\le \sqrt{n}(t-\kappa_0^\eto))- G^\ieo_{n,q,\sqrt{n}(t-\kappa_0^\ieo)}- G_{n,q}^\eto
\end{align*}

We further discuss the case when (2) $\kappa_0^\ieo<t<\kappa_0^\eto$, (3) $t=\kappa_0^\eto$, (4) $t>\kappa_0^\eto$.

When $\kappa_0^\ieo<t<\kappa_0^\eto$, we have $\PP(\nabla_\thete v_0(\omege_{\thete^\kl})\NN_1^\eto\le \sqrt{n}(t-\kappa_0^\eto))\leq \exp\rbr{-\frac{n(\kappa_0^\eto - t)^2}{2\nbr{\nabla_\thete v_0(\omege_{\thete^\kl})\bM_1^\eto}^2}}$ by Lemma \ref{lemma: tailchisquare}.
So
\begin{align*}
&\PP(R(\hat{\omege}^\eto)\geq t) - \PP(R(\hat{\omege}^\ieo)\geq t) \\
\ge & 1 - \exp\rbr{-\frac{n(\kappa_0^\eto - t)^2}{2\nbr{\nabla_\thete v_0(\omege_{\thete^\kl})\bM_1^\eto}^2}}- G^\ieo_{n,q,\sqrt{n}(t-\kappa_0^\ieo)}- G_{n,q}^\eto
\end{align*}

When $t=\kappa_0^\eto$, we have $\PP(\nabla_\thete v_0(\omege_{\thete^\kl})\NN_1^\eto\le \sqrt{n}(t-\kappa_0^\eto))=\frac{1}{2}$. So
$$\PP(R(\hat{\omege}^\eto)\geq t)-\PP(R(\hat{\omege}^\ieo)\geq t)  \ge \frac{1}{2} - G^\ieo_{n,q,\sqrt{n}(t-\kappa_0^\ieo)}- G_{n,q}^\eto$$

When $t > \kappa_0^\eto$, we have $1-\PP(\nabla_\thete v_0(\omege_{\thete^\kl})\NN_1^\eto\le \sqrt{n}(t-\kappa_0^\eto))\ge 0$. 
So
\begin{align*}
& \PP(R(\hat{\omege}^\eto)\geq t)-\PP(R(\hat{\omege}^\ieo)\geq t) \\
\ge & -G^\ieo_{n,q,\sqrt{n}(t-\kappa_0^\ieo)}- G_{n,q}^\eto.
\end{align*}
\endproof


\proof[Proof of Theorem \ref{thm: finitecomparemis1}]
In the following analysis, we define $t':=t-\kappa_0^\ieo$.

Case 1: $t\le \kappa_0^{\ieo}$. In this case, $R(\hat{\omege}^{\ieo})\ge R(\omege_{\thete^*}) = \kappa_0^{\ieo}\ge t$ and $R(\hat{\omege}^{\eto})\ge R(\omege_{\thete^*}) = \kappa_0^{\ieo}\ge t$ as any realization of $\hat{\omege}^{\ieo}$ or $\hat{\omege}^{\eto}$ should have larger regret than $\omege_{\thete^*}$. So
$$\PP(R(\hat{\omege}^{\eto})\ge t) - \PP( R(\hat{\omege}^{\ieo})\ge t)=1-1=0.$$

Case 2: {$\delta>0$, $\tau_1>0$, and} $t> \kappa_0^{\ieo}+\frac{\tau_6+\tau_1}{\tau_1} \delta$, i.e., $t' > \frac{\tau_6+\tau_1}{\tau_1} \delta$. In this case, we have that $(1+\frac{\tau_1}{\tau_6})(nt'-n\delta) > nt'$. Suppose $\delta>0$ and consider any $0< \varepsilon < {\frac{\tau_1}{\tau_6+\tau_1}} t' - \delta$.

We first observe that the Berry-Esseen bound implies the following inequality. Since $\nabla_\thete v_0(\omege_{\thete^\kl})(\cdot)$ is an affine function:
\begin{align*}
    \sup_{t\geq 0}|\PP(\sqrt{n}\nabla_{\thete}v_0(\omege_{\thete^\kl})(\hat{\thete}^\eto-\thete^\kl)\geq t)-\PP(\nabla_\thete v_0(\omege_{\thete^\kl})\NN_1^\eto\geq t)|\leq C_{n,q}^\eto. 
\end{align*}

For any sample size $n$ and any given $\varepsilon>0$, we have
\begin{align*}
& \PP(R(\hat{\omege}^{\eto})\ge t) - \PP( R(\hat{\omege}^{\ieo})\ge t) \\
= & \PP(n R(\hat{\omege}^{\eto})\ge nt) - \PP(n R(\hat{\omege}^{\ieo})\ge nt) \\
= & \PP(n (R(\hat{\omege}^{\eto})-R(\omege_{\thete^{\kl}}))\ge nt-n\kappa_0^{\eto}) - \PP(n (R(\hat{\omege}^{\ieo})-R(\omege_{\thete^{*}}))\ge nt-n\kappa_0^{\ieo}) \\
= & \PP(n (R(\hat{\omege}^{\eto})-R(\omege_{\thete^{\kl}}))\ge nt'-n\delta) - \PP(n (R(\hat{\omege}^{\ieo})-R(\omege_{\thete^{*}}))\ge nt')\\
 & \quad \quad (\text{where } \delta=\kappa_0^{\eto}-\kappa_0^{\ieo}\ge 0 \ t':=t-\kappa_0^{\ieo})\\
\le & \PP(n\rbr{R(\hat{\omege}^{\eto})-\kappa_0^{\eto}-\nabla_{\thete}\vale_0(\omege_{\thete^{\kl}})(\hat{\thete}^{\eto}-\thete^{\kl})}\ge nt'-n\delta-n\varepsilon)  \\
&+ {\PP(n\nabla_{\thete}\vale_0(\omege_{\thete^{\kl}})(\hat{\thete}^{\eto}-\thete^{\kl})\ge n\varepsilon)} \\
&
- \PP(n (R(\hat{\omege}^{\ieo})-R(\omege_{\thete^{*}}))\ge nt')\\
\le & \PP(\GG^{\eto} \ge nt'-n\delta-n\varepsilon) + \PP(\nabla_{\thete}\vale_0(\omege_{\thete^{\kl}}) {\bM_1^{\eto}} \bm{Y}_0\ge \sqrt{n}\varepsilon) \\
&-  \PP(\GG^{\ieo} \ge nt') +  D_{n,q}^{\eto}+D_{n,q}^{\ieo} + C_{n,q}^{\eto} \\
\le & \PP(\GG^{\eto} \ge nt'-n\delta-n\varepsilon)
-  \PP(\GG^{\ieo} \ge nt')+  D_{n,q}^{\eto}+D_{n,q}^{\ieo} + C_{n,q}^{\eto} +  E_{n}^{\delta, \varepsilon}
\end{align*}    
where, by Lemma \ref{lemma: tailchisquare},
$$E_{n}^{\delta, \varepsilon} = \exp{\left(-\frac{n \varepsilon^2}{2\|\nabla_{\thete}\vale_0(\omege_{\thete^{\kl}}) {{\bM}_1^{\eto}}\|^2}\right)}.$$
Therefore, we only need to provide the bounds for 
$$\PP(\GG^{\eto} \ge nt'-n\delta-n\varepsilon)
-  \PP(\GG^{\ieo} \ge nt').$$
To achieve this, we can apply Theorem \ref{thm: Ginmis} to claim that for all $s\geq 0$, 
$$\PP({\mathbb G}^{\eto} \ge s)\leq \PP({\mathbb G}^{\ieo} \ge (1+\frac{\tau_1}{\tau_6})s)\leq\PP({\mathbb G}^{\ieo} \ge s).$$ To see this, we note that
\begin{align*}
& \PP(\GG^{\eto} \ge s) \\ 
=& \PP\Big(\frac{1}{2} \bm{Y}_0^\top {\bD^{\eto}} \bm{Y}_0 \ge s \Big) \\
=& \PP\Big(\frac{1}{2} \bm{Y}_0^\top {\bD^{\eto}} \bm{Y}_0 \ge s \text{ and } \frac{1}{2} \tau_6 \bm{Y}_0^\top \bm{Y}_0 \ge s \Big)\tag{since $\tau_6$ is the largest eigenvalue of ${\bD^{\eto}}$}\\
=& \PP\Big(\frac{1}{2} \bm{Y}_0^\top {\bD^{\eto}} \bm{Y}_0 \ge s \text{ and } \frac{1}{2} \tau_1 \bm{Y}_0^\top \bm{Y}_0 \ge  \frac{\tau_1}{\tau_6}s \Big)\\
=& \PP\Big(\frac{1}{2} \bm{Y}_0^\top {\bD^{\eto}} \bm{Y}_0 \ge s\\
& \text{ and } \frac{1}{2} \bm{Y}_0^\top (
{\bD^{\ieo}}  - {\bD^{\eto}}
) \bm{Y}_0 \ge  \frac{\tau_1}{\tau_6}s \Big)\tag{since $\tau_1$ is the smallest eigenvalue of $(
{\bD^{\ieo}}  - {\bD^{\eto}}
)$}\\
\le & \PP\Big(\frac{1}{2} \bm{Y}_0^\top
{\bD^{\ieo}}  \bm{Y}_0 \ge  (1+\frac{\tau_1}{\tau_6})s \Big)\\
= & \PP({\mathbb G}^{\ieo} \ge (1+\frac{\tau_1}{\tau_6})s).
\end{align*}


This shows that
\begin{align*}
&\PP(\GG^{\eto} \ge nt'-n\delta-n\varepsilon)
-  \PP(\GG^{\ieo} \ge nt')\\
\le & \PP(\GG^{\ieo} \ge (1+\frac{\tau_1}{\tau_6})(nt'-n\delta-n\varepsilon))-\PP(\GG^{\ieo} \ge nt')\\
= &
-\PP(nt' \le \GG^{\ieo} \le (1+\frac{\tau_1}{\tau_6})(nt' -n\delta-n\varepsilon))
\end{align*}
In conclusion, 
\begin{align*}
    &\PP(R(\hat{\omege}^{\eto})\ge t) - \PP( R(\hat{\omege}^{\ieo})\ge t)\\
    \leq  & \PP(\GG^{\eto} \ge nt-n\kappa_0^\ieo-n\delta-n\varepsilon)
-  \PP(\GG^{\ieo} \ge nt-n\kappa_0^\ieo) + D_{n,q}^{\eto}+D_{n,q}^{\ieo} + C_{n,q}^{\eto} +  E_{n}^{\delta, \varepsilon}\\
\le & -\PP(nt-n\kappa_0^\ieo \le \GG^{\ieo} \le (1+\frac{\tau_1}{\tau_6})(nt- n\kappa_0^\ieo -n\delta-n\varepsilon)) + D_{n,q}^{\eto}+D_{n,q}^{\ieo} + C_{n,q}^{\eto} +  E_{n}^{\delta, \varepsilon}.
\end{align*}

Finally, we remark that when $\delta=0$, $\theta^*=\theta^\kl$, $\nabla_{\thete}\vale_0(\omege_{\thete^{\kl}})=0$, and $\kappa_0^{\eto}=\kappa_0^{\ieo}$. In this case, there is no need to introduce $\varepsilon$, and we have similarly
\begin{align*}
& \PP(R(\hat{\omege}^{\eto})\ge t) - \PP( R(\hat{\omege}^{\ieo})\ge t) \\
= & \PP(n(R(\hat{\omege}^{\eto})-R(\omege_{\thete^{*}}))\ge nt')
- \PP(n (R(\hat{\omege}^{\ieo})-R(\omege_{\thete^{*}}))\ge nt')\\
\le & \PP(\GG^{\eto} \ge nt')-  \PP(\GG^{\ieo} \ge nt') +  D_{n,q}^{\eto}+D_{n,q}^{\ieo}\\
\le & \PP(\GG^{\ieo} \ge (1+\frac{\tau_1}{\tau_6}) nt')-  \PP(\GG^{\ieo} \ge nt') +  D_{n,q}^{\eto}+D_{n,q}^{\ieo}\\
= & - \PP(nt'\le \GG^{\ieo} \le (1+\frac{\tau_1}{\tau_6}) nt') +  D_{n,q}^{\eto}+D_{n,q}^{\ieo}.
\end{align*}

\endproof

\end{document}